\documentclass[runningheads]{llncs}

\usepackage[mobile]{eccv}

\usepackage{eccvabbrv}

\usepackage{graphicx}
\usepackage{booktabs}

\usepackage[accsupp]{axessibility}  

\usepackage[breaklinks,colorlinks,citecolor=eccvblue]{hyperref}

\usepackage{orcidlink}

\newcommand{\new}[1]{{\color{black}{#1}}}

\usepackage{fancyhdr}
\fancypagestyle{firstpage}{
  \fancyhf{}
  \fancyhead[C]{Paper preprint version accepted at\\ 2026 European Conference on Computer Vision (ECCV) \vspace{0.05cm}} 
}

\begin{document}

\title{SVI360: Spherical Video Interpolation}
\titlerunning{SVI360: Spherical Video Interpolation}
\date{\today}

\author{Le-Kim Nguyen\inst{1}\orcidlink{0009-0005-1572-636X} \and
Renato Martins\inst{1}\orcidlink{0000-0003-0053-0004} \and\\
Pascal Vasseur\inst{2}\orcidlink{0000-0001-5145-9653} \and
Cedric Demonceaux\inst{1}\orcidlink{0000-0001-6916-1273}}

\authorrunning{Nguyen et al.}

\institute{Université Bourgogne Europe, ICB UMR 6303 CNRS, France\\
\and
Université de Picardie Jules Verne, MIS UR 4290, France}

\maketitle

\begin{abstract}
This paper addresses the problem of omnidirectional video interpolation,
which plays an essential role in applications such as virtual reality
and immersive video enhancement. Existing video interpolation methods
are not well-suited for spherical videos, as they have difficulty
handling severe distortions close to the poles. To address this issue, we propose SVI360, a dual-branch framework
that combines the image frame and its rotated orthogonal view to deal with these distortions. The core methodological aspect of the approach is to reinforce equivariance of the flow displacements between the original and orthogonal views to improve intermediate frame prediction.  
Experiments show that
our method outperforms state-of-the-art approaches in interpolation
quality while maintaining accurate optical flow in four different public benchmarks. \new{Code and pre-trained models are available at:}
\url{https://icb-vision-ai.github.io/video360_interpolation/}
\keywords{Video frame interpolation \and Omnidirectional images \and Spherical optical flow estimation.  }
\end{abstract}

\thispagestyle{firstpage}

\section{Introduction}
\begin{figure}[t]
  \centering
  \includegraphics[width=1\linewidth]{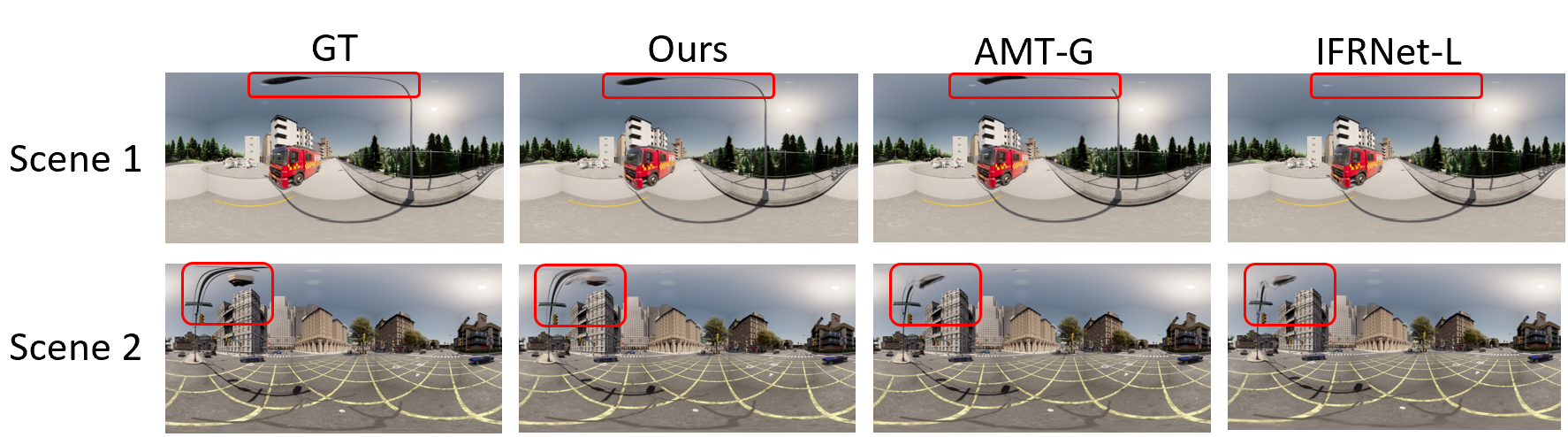}
  \caption{Limitations of perspective video interpolation methods on spherical images. Red boxes indicate regions where existing methods produce severe ghosting and blurring, whereas the proposed approach better preserves fine image details.}
  \label{fig:sota_limitation}
\end{figure}

Spherical video, also known as omnidirectional or panoramic video,
has attracted increasing research interest and applicability since it encapsulates 360-scene information for each image frame. This is a key capability in virtual reality, navigation and robotics. As their application increases across domains such as entertainment, education, and
training, the demand for high-quality 360-degree videos becomes even
more critical. In this context, achieving smooth video frame transitions is particularly important, as
higher frame rates greatly enhance user immersion and help reduce
cybersickness. However, streaming or transmitting high frame rate 360$^{\circ}$ videos requires
substantial bandwidth~\cite{10.1145/3083187.3083190}, making it inefficient in many real-world
settings. To mitigate this issue, video interpolation has emerged as a promising solution, 
enabling the synthesis of intermediate frames between two consecutive images. By generating additional
 frames locally, such methods can deliver smoother frame transitions while significantly reducing transmission 
 requirements. Existing approaches~\cite{superslomo,film,ifrnet,vfiformer,amt} have achieved
  impressive results on perspective images. Yet their performance drops sharply when applied to spherical 
imagery (please see some examples in~\cref{fig:sota_limitation}). \\
\indent Compared with typical perspective images, spherical data introduce severe geometric distortions, non-uniform pixel
 density, and large pixel flow displacements. These characteristics make
  correspondence estimation and motion reconstruction considerably more challenging, emphasizing the need 
  for architectures designed to handle spherical data.  
Despite the growing importance of 360$^{\circ}$ video, omnidirectional video interpolation methods 
remain clearly overlooked. 
In this context, 360VFI\cite{360vfi} is the first
method and benchmark dedicated to 360$^{\circ}$ frame interpolation. It introduces a
distortion-aware design that uses spherical geometry priors for feature extraction and
fusion, and provides an evaluation protocol based on flow magnitude. They stratified the test set into four distinct settings based on the vertical flow magnitude, where the flow may result from object motion and camera movement. In the Easy settings, characterized by very small pixel displacement between two frames (average flow magnitude <2), the method can perform well. However, as the flow magnitude increases, the performance drops significantly, indicating that the approach still has limitations in handling large pixel displacements.
In this work, we introduce
SVI360, a novel video interpolation approach designed to improve the fidelity of interpolated
frames while maintaining high-quality optical flow estimation, particularly in image regions containing larger distortions. 

Our method adopts a coarse-to-fine base
strategy with progressively increasing resolutions, inspired by AMT~\cite{amt}, where intermediate optical
flows and intermediate frame features are iteratively refined using dense
all-pairs bidirectional correlation volumes.
To adapt this concept to spherical geometry, we incorporate the key hypothesis
from PriOr-Flow\cite{priorflow}, which suggests that generating orthogonal views
of spherical images can provide complementary priors that improve optical flow
estimation in the primitive (original) spherical view. Based on this insight, we design a
dual-branch architecture where the orthogonal branch refines both the optical
flow and the interpolated frame in the primitive branch through cross-view
feature interaction, yielding more accurate frame interpolations in four widely adopted public datasets: two synthetic datasets (FlowScape~\cite{flowscape} and Flow360~\cite{flow360}) and two
real-world datasets (ODV360~\cite{odv360} and 360VFI~\cite{360vfi}). 
The obtained results consistently demonstrate the capability of addressing distortion near the two poles while also maintaining strong performance under large motion fields.
On FlowScape, where
the task is middle-frame interpolation, SVI360 surpasses the best competitor by
+0.63\,dB in PSNR and +0.33\,dB in WS-PSNR, while also achieving higher accuracy for optical flow estimation. On Flow360, which evaluates interpolation at
arbitrary temporal positions, SVI360 also has the best performance on all reported
interpolation metrics, including +0.25\,dB PSNR gain over the second-best method.
Clear improvements are also observed when considering real-world ODV360 and 360VFI benchmarks, indicating the learned interpolation model also presents good generalization capabilities across scenes.
Visual inspection on frames from several scenes also agrees with these findings.
To increase interpolation accuracy while maintaining competitive computational cost, our work brings three main contributions:
\begin{enumerate}
    \item We present a new omnidirectional video interpolation method that leverages equivariance pixel displacement priors from a rotated orthogonal view of the sphere to guide optical flow estimation and frame interpolation of the original spherical frame.
    \item In order to account for non-uniform pixel density along the spherical view, we introduce a Spherical Weighted Charbonnier reconstruction loss term, enhancing performance in the central image regions that contain the majority of visual information.
    \item Extensive evaluations on omnidirectional benchmarks, including FlowScape, Flow360, ODV360, and 360VFI are provided, where our method consistently outperforms state-of-the-art approaches in terms of interpolation quality while achieving competitive optical flow accuracy.\end{enumerate}
\section{Related Work}
\label{sec:related}

\subsection{Perspective Video Frame Interpolation}
Video frame interpolation (VFI) for perspective images has been extensively studied and 
can be broadly categorized into phase-based, kernel-based, motion-based\new{, and more recently generative}  approaches~\cite{phaseinterp,phasenet,adaconv,sepconv,adacof, FCVG}. 
Phase-based methods~\cite{phaseinterp,phasenet} represent motion as pixel-wise phase shifts and reconstruct 
intermediate frames by interpolating phase and amplitude across a multi-scale 
pyramid, 
but they are often limited to small pixel displacements. On the other hand, kernel-based approaches such as AdaCoF~\cite{adacof} further adopt deformable convolutions to adjust the receptive field. 
  However, these methods often yield blurry results for large pixel displacements and require large kernels 
  with many parameters to handle large displacements. Motion-based VFI approaches have achieved substantial improvements with the recent advances of optical flow methods~\cite{superslomo, rife,vfiformer}. Recent methods adopt a coarse-to-fine refinement strategy \cite{film,ifrnet,amt}. In particular, AMT~\cite{amt} reconstructs dense all-pairs correlation volumes to model large-displacement correspondences, which inspires our method. \new{In parallel, recent generative approaches have explored video inbetweening by leveraging diffusion-based video generation models. For instance, FCVG~\cite{FCVG} introduces a frame-wise condition to guide video generation to improve temporal stability under complex motions.} Despite their strong performance on perspective images, these methods degrade considerably on omnidirectional images due to their reliance on planar projection assumptions which limits their direct applicability to omnidirectional data. In contrast, our approach presents a dual-branch AMT base model, with carefully designed major adaptations to handle larger pixel displacements and distortions of omnidirectional images.

\subsection{Spherical Video Interpolation and Optical Flow}
Processing omnidirectional images requires the adaptation of typical perspective signal operations~\cite{spherenet,ktn,fernandezlabrador2019cornerslayoutendtoendlayout}. SphereNet~\cite{spherenet} and EquiConvs\cite{fernandezlabrador2019cornerslayoutendtoendlayout} propose geometry-aware spherical convolutions that account for distortion and rotational properties of the sphere. Building on a similar idea, Spherical CNNs~\cite{sphericalcnn} perform convolution directly on the sphere to achieve rotation-equivariant representations. Another trend is to reduce distortion by splitting spherical images into multiple perspective views. Features are extracted from tangent-plane projections and fused across views \cite{sphericalconvfast}; cubemap-based methods follow the same strategy but often suffer from discontinuities at view boundaries and increased system complexity. These operations have been transposed to optical flow estimation, \eg SLOF \cite{flow360} and PriOr-Flow \cite{priorflow}, mitigating polar distortions by exploiting cross-view consistency. However, frame interpolation to omnidirectional imagery remains underexplored with 360VFI~\cite{360vfi} being the only representative method. 360VFI mitigates spherical distortions by incorporating distortion-aware priors in both feature extraction and frame synthesis. Nevertheless, its learned offset design from distortion maps is less well justified than static precomputed offsets such as EquiConvs, which are effective in practice, and it does not explicitly handle large pixel displacements. Our work addresses these limitations through a coarse-to-fine strategy and spherical refinement, resulting in better interpolations consistently even for increasing flow displacements as shown in the ``Hard''  and ``Extreme'' splits in the experiments. 

\section{Method}
\label{sec:method}
\begin{figure}[t]
  \centering
  \includegraphics[width=\linewidth]{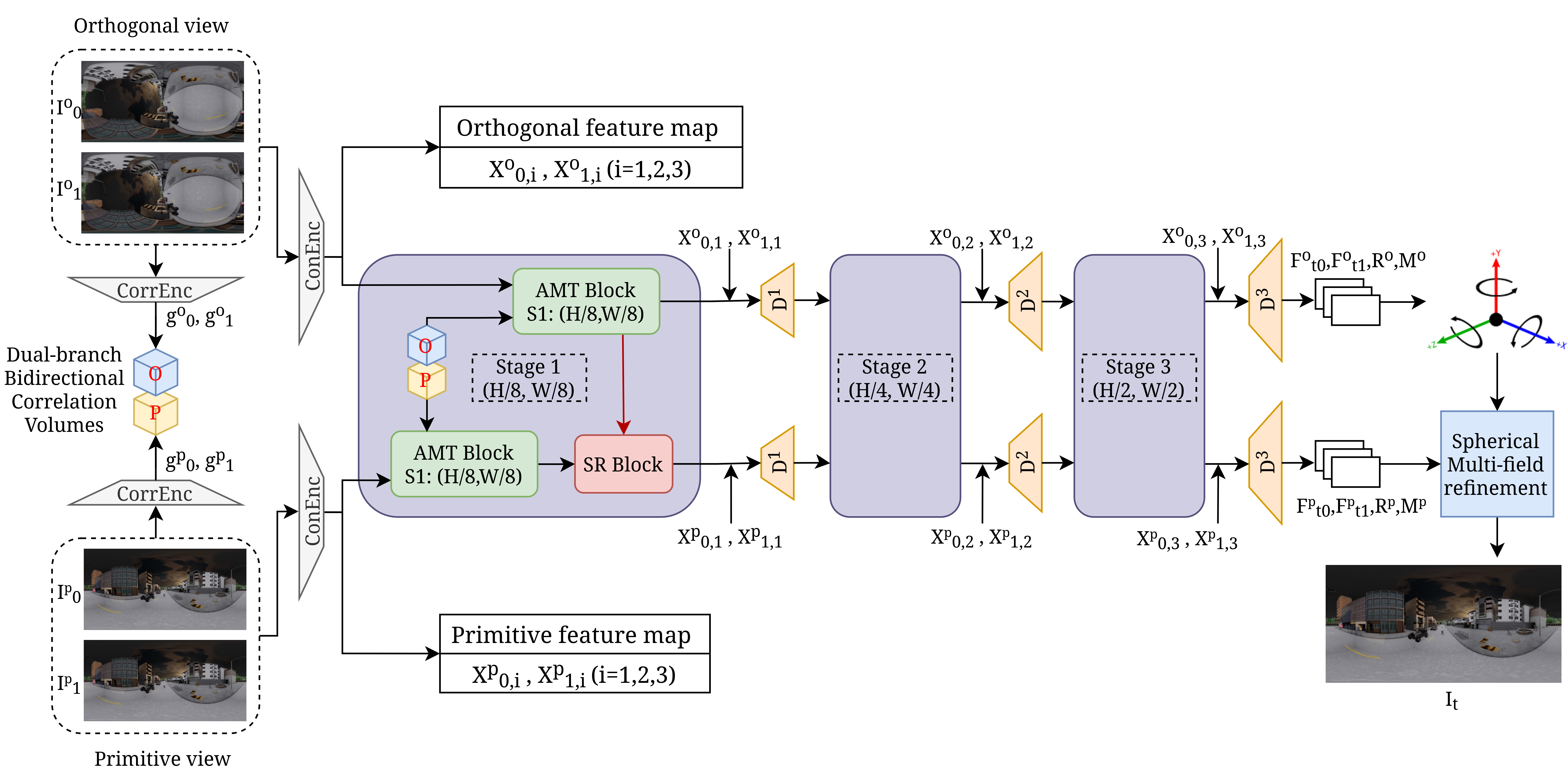}
  \caption{Overview of SVI360. The primitive (original) spherical frame and its rotated orthogonal view are processed by separate encoders, which are then reconstructed with a three-stage coarse-to-fine decoder. Correlation volumes refine optical flow and intermediate features across stages, while the ``Spherical Refiner'' component transfers orthogonal cues to the primitive branch to correct distortions. Finally, the spherical multi-field refinement component fuses the outputs from the two branches to produce the final interpolated frame.}
  \label{fig:proposed_method}
\end{figure}

Given two input spherical frames $\textbf{I}_0$ and $\textbf{I}_1$, our goal is to predict the intermediate frame $\textbf{I}_t$ at time $t\in(0,1)$. An overview of our approach is given in Fig.~\ref{fig:proposed_method}. We first generate an orthogonal view for each input frame by applying a $90^\circ$ rotation around the x-axis. This design follows the distortion characteristics of equirectangular images: equatorial regions in primitive view provide more reliable motion cues, while polar regions are heavily distorted. After rotation, the original polar regions move closer to the center in the orthogonal view allowing more accurate flow estimation in these regions\cite{priorflow}. \new{Once the orthogonal view is processed, its motion information is rotated back to the original orientation and used to refine the primitive branch. Unlike PriOr-Flow, which focuses on optical flow refinement, SVI360 is designed for spherical video frame interpolation. It performs multi-stage coarse-to-fine refinement, jointly updating optical flow and intermediate features to improve the final interpolated frame}. We will describe the details of each component in the following sections.

\subsection{Image Rotation}
The input \new{equirectangular} frame $\textbf{I}(u,v)$ is first mapped
from 2D pixel coordinates to spherical coordinates $(\theta,\phi)$, and then converted
to the corresponding 3D Cartesian coordinate $(x,y,z)$. PriOr-Flow~\cite{priorflow} found that empirically a rotation angle of $\alpha=90^\circ$
around the x-axis optimally reduces polar distortions, allowing more accurate optical flow
estimation in high-latitude regions. Finally, the rotated 3D coordinates are projected back
to the equirectangular \new{representation} to obtain the orthogonal view. We denote this transformation from the primitive view to the orthogonal view as $\textbf{T}_{p}^{o}$,
and its inverse as $\textbf{T}_{o}^{p}$. Given two consecutive equirectangular frames
$\{\textbf{I}_0^p, \textbf{I}_1^p\}$, the corresponding orthogonal views are obtained as:
\begin{equation}
\{\textbf{I}_0^o, \textbf{I}_1^o\}
=
\{\textbf{T}_{p}^{o}(\textbf{I}_0^p),\,\textbf{T}_{p}^{o}(\textbf{I}_1^p)\}.
\end{equation}

\subsection{Correlation Volume Construction and Lookup}
\begin{figure}[t]
  \centering
  \begin{minipage}[t]{0.45\linewidth}
    \centering
    \includegraphics[width=\linewidth]{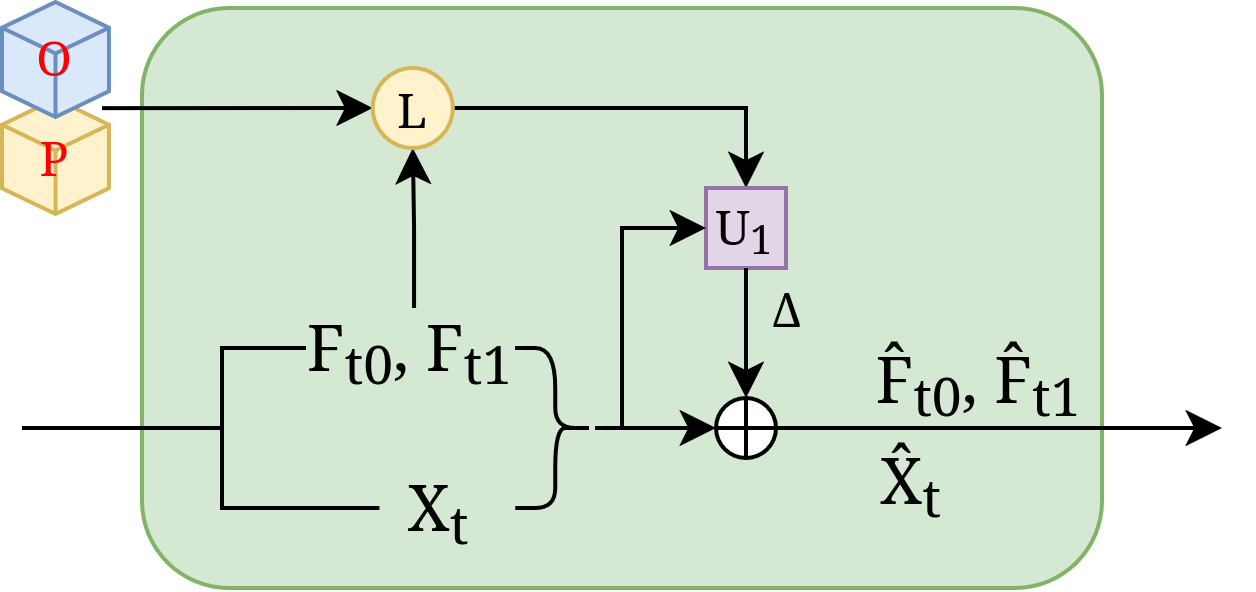}

    {\footnotesize (a) AMT-style update.}
  \end{minipage}
  \hspace{1cm}
  \begin{minipage}[t]{0.3\linewidth}
    \centering
    \includegraphics[width=\linewidth]{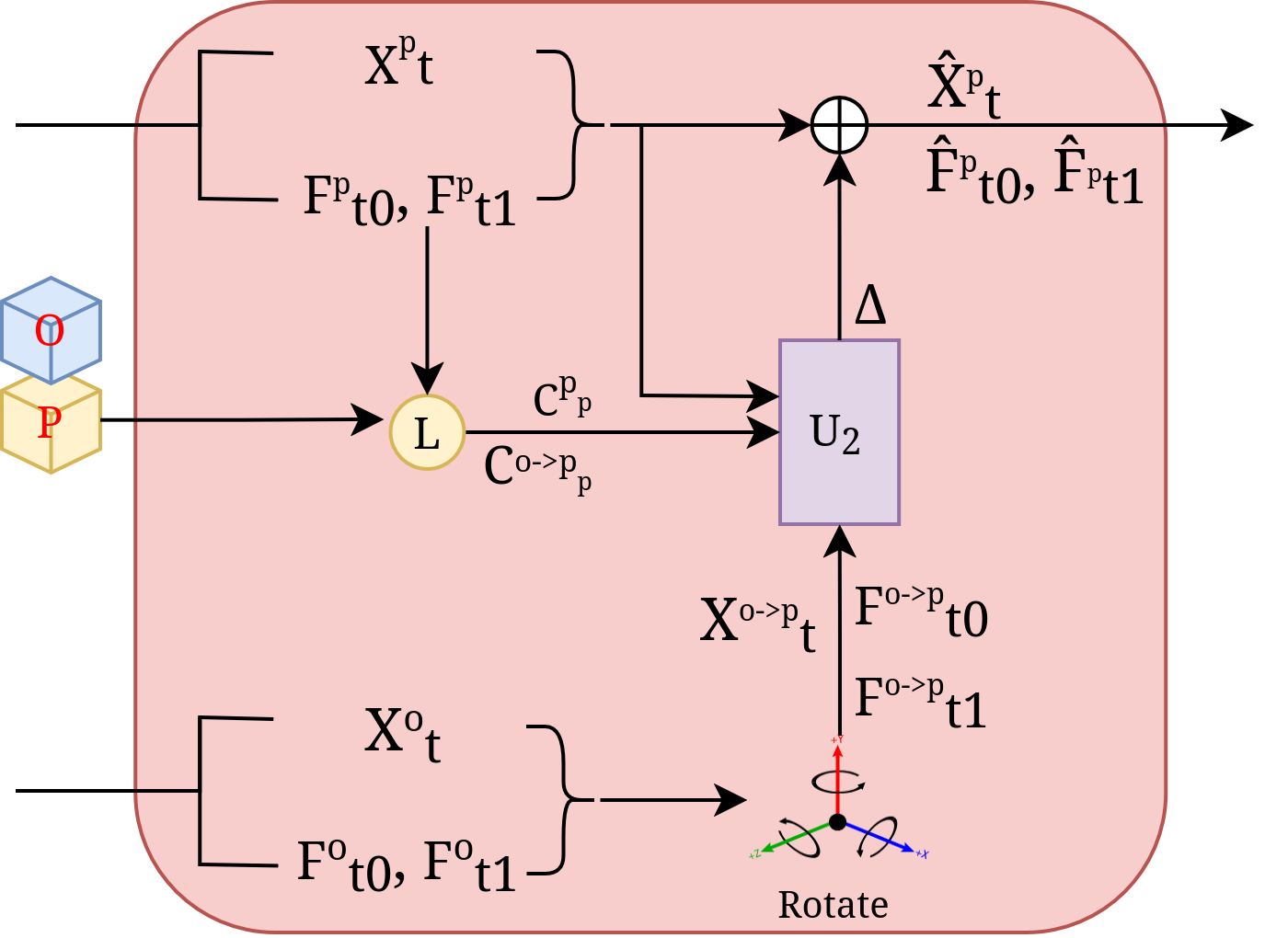}

    {\footnotesize (b) Spherical refiner.}
  \end{minipage}
  \caption{Update modules in SVI360. (a) AMT-style recurrent update used in both primitive and orthogonal branches to refine bidirectional flow and intermediate features with DCCL cues. (b) Spherical Refiner component that fuses cross-view information. $\mathcal{L}$ denotes the correlation lookup operator that retrieves correlation features based on the current flow estimate. $\mathcal{U}_1$ and $\mathcal{U}_2$ are update networks that predict residual corrections for flow and intermediate features; Further details on the architecture are in the Supplementary Material.}
  \label{fig:update_modules}
\end{figure}

\new{
Following RAFT~\cite{raft}, we build an all-pairs correlation volume for each branch $b\in\{p, o\}$, where $p$, $o$ denotes the primitive and orthogonal branch respectively. Given the feature maps $\mathbf{g}_0^b$ and $\mathbf{g}_1^b$ extracted from the two input frames in the corresponding branch $b$, the correlation volume is computed as: \begin{equation} \mathcal{C}^{b}_{ijkl} = \sum_h (\mathbf{g}_0^{b})_{hij} (\mathbf{g}_1^{b})_{hkl}. \end{equation} We further construct a 4-level correlation pyramid $\{\mathcal{C}_i^{b}\}_{i=1}^{4}$ via average pooling along the last two dimensions.
For correlation lookup, we adopt the Dual-Cost Collaborative Lookup (DCCL) strategy from PriOr-Flow~\cite{priorflow}. 
Given the current flow estimate, the lookup operation retrieves two types of correlation cues: one from the current branch and one from the corresponding rotated branch. 
We denote this operation as
\begin{equation}
\mathbf{C}^{b}, \mathbf{C}^{\bar{b}\rightarrow b}
=
\operatorname{DCCL}
\left(
\{\mathcal{C}_i^{b}\}_{i=1}^{4},
\{\mathcal{C}_i^{\bar{b}}\}_{i=1}^{4},
\mathbf{f}^{b}
\right),
\end{equation}
where $b\in\{p,~o\}$ is the current branch, $\bar{b}$ is the counterpart branch, and $\mathbf{f}^{b}$ is the estimated flow of current branch.
Intuitively, DCCL first uses the current flow to locate matching regions between two primitive views. 
Then, it also checks the corresponding regions in the rotated view, where some distorted areas may become easier to match. The resulting same-view and cross-view cues are jointly fed into the AMT-style update component (Fig.~\ref{fig:update_modules}(a)), {enabling the refinement of both optical flow and intermediate features, instead of only refining the optical flow as in PriOr-Flow}. Specifically, following the update component as in AMT~\cite{amt}, each branch applies a lightweight recurrent update block that takes the current flow, correlation features, and intermediate features as input, and predicts residuals for both optical flow and intermediate features. The use of these cues in the Spherical Refiner is described in the next section.
}

\subsection{Spherical Refiner}
After updating the optical flow and feature maps in both branches, we feed them into the Spherical Refiner. This component is designed as a residual update that jointly refines the optical flow and intermediate representations solely for the primitive branch by taking as input the primitive branch cues ($\mathbf{f}^p, \mathbf{X}_t^p$) and the orthogonal branch cues ($\mathbf{f}^o, \mathbf{X}_t^o$). 
To align both views in a common reference frame, we first rotate the orthogonal flow and intermediate features back to the primitive domain:
\begin{equation}
\mathbf{f}^{o \rightarrow p} = \textbf{T}_{o}^{p}(\mathbf{f}^o), 
\qquad
\mathbf{X}_t^{o \rightarrow p} = \textbf{T}_{o}^{p}(\mathbf{X}_t^{o}),
\end{equation}
Given the updated flow estimates, we perform correlation lookup again to obtain the primitive correlation feature maps $\mathbf{\textbf{C}}^p$ and $\mathbf{\textbf{C}}^{o \rightarrow p}$. 
All cues are then fed into the update block $\mathcal{U}_2$, as illustrated in Fig.~\ref{fig:update_modules}(b), to predict residual corrections for the primitive branch:
\begin{equation}
\Delta \mathbf{f}^p, \Delta \mathbf{X}_t^p
=
\mathcal{U}_2\!\left(
\mathbf{f}^p,\,
\mathbf{X}_t^p,\,
\mathbf{C}^p,\,
\mathbf{C}^{o \rightarrow p},\,
\mathbf{X}_t^{o \rightarrow p},\,
\mathbf{f}^{o \rightarrow p}
\right).
\end{equation}
The refined primitive flow and feature maps are then updated as:
\begin{equation}
\mathbf{f}^p \leftarrow \mathbf{f}^p + \Delta \mathbf{f}^p, 
\qquad
\mathbf{X}_t^p \leftarrow \mathbf{X}_t^p + \Delta \mathbf{X}_t^p.
\end{equation}
\new{Then, we upsample the flows and intermediate features using the AMT decoder design~\cite{amt}, but applied symmetrically to both the primitive and orthogonal branches. These upsampled representations are then passed to the next stage, where refinement is repeated at a higher resolution. This coarse-to-fine design allows the model to first estimate large motions at low resolution, where displacements are easier to track, and then progressively refine local details at higher resolutions.}

\subsection{Spherical Multi-field Refinement}
Flow-based VFI methods synthesize the intermediate frame by backward warping the input frames using bilateral flows, combined with an occlusion mask and a residual term:
\begin{equation}
\textbf{I}_t = \textbf{M} \odot \mathcal{W}(\textbf{I}_0, \textbf{F}_{t\rightarrow0}) + (1-\textbf{M}) \odot \mathcal{W}(\textbf{I}_1, \textbf{F}_{t\rightarrow1}) + \textbf{R}.
\end{equation}
where $\textbf{M}$ is the occlusion mask, $\textbf{R}$ is the residual term, $\mathcal{W}$ is backward warping operation, and $\odot$ denotes element-wise multiplication. Predicting only a single pair of flows limits the possible solutions in occluded or highly distorted regions \cite{amt}. \new{Therefore, the last decoder of each branch does not output only one flow field, occlusion mask, and residual term. 
Instead, it predicts $N$ candidate sets of flow fields, occlusion masks and residuals. 
Each candidate set is then used in the frame synthesis formulation above to produce one intermediate frame candidate. In our implementation, we set $N=5$, so each branch produces five intermediate frame candidates. {Different from PriOr-Flow, where the orthogonal view is mainly used to refine solely the flow, our orthogonal branch also contributes directly to image synthesis.}}
The primitive and orthogonal branches focus on different distortion characteristics of panoramic imagery, so their candidate frames provide complementary cues.
In the orthogonal branch, the network predicts $\{\textbf{I}_{t}^{o,1},\ldots,\textbf{I}_{t}^{o,N}\}$.
These frames are then rotated back to the primitive spherical view:
\begin{equation}
\textbf{I}_{t}^{o2p,n}=\textbf{T}_{o}^{p}\!\left(\textbf{I}_{t}^{o,n}\right), \quad n=1,\ldots,N.
\end{equation}
After alignment, the primitive candidates
$\{\textbf{I}_{t}^{p,1},\ldots,\textbf{I}_{t}^{p,N}\}$ and the rotated orthogonal candidates
$\{\textbf{I}_{t}^{o2p,1},\ldots,\textbf{I}_{t}^{o2p,N}\}$ are jointly fused to produce the final
interpolated frame:
\begin{equation}\label{eq:fuse}
\textbf{I}_t=\mathcal{F}_{\text{fuse}}\!\left(
\{\textbf{I}_{t}^{p,1},\ldots,\textbf{I}_{t}^{p,N}\},
\{\textbf{I}_{t}^{o2p,1},\ldots,\textbf{I}_{t}^{o2p,N}\}
\right),
\end{equation}
where $\textbf{I}_{t}^{p,n}$ denotes the $n$-th candidate from the primitive branch and
$\textbf{I}_{t}^{o2p,n}$ denotes the corresponding candidate from the orthogonal branch after
inverse rotation. $\mathcal{F}_{\text{fuse}}(\cdot)$ is implemented as a lightweight
two-layer convolutional fusion head. This refinement aggregates complementary information across both spherical views.

\subsection{Loss Functions}
We supervise intermediate frame synthesis with a combination of two reconstruction error terms: census~\cite{meister2017unflowunsupervisedlearningoptical} and Charbonnier~\cite{413553}:
\begin{equation}
\mathcal{L}_{r}=\mathcal{L}_{\text{cen}} + \lambda\mathcal{L}_{\text{char}}.
\end{equation}
For structure preservation under illumination changes, $\mathcal{L}_{\text{cen}}$
\cite{meister2017unflowunsupervisedlearningoptical} computes a soft Hamming distance
between census-transformed patches (size $7\times 7$) of
$\hat{\textbf{I}}_t$ and $\textbf{I}_t^{gt}$.
We also adapt a Charbonnier reconstruction term because it is a smooth, robust alternative to $\ell_1$ and
is less sensitive to outliers in occluded and highly distorted regions.
In the original Charbonnier formulation\cite{413553}, all pixels contribute equally to the loss. However, for spherical images, distortions near the poles often lead to significantly larger errors compared to equatorial regions. Consequently, the loss can be dominated by polar regions, causing the model to prioritize minimizing polar errors while underemphasizing central regions that contain the majority of visual information. To mitigate this effect, we propose a spherical-weighted Charbonnier loss with a latitude-dependent weight defined as:
\begin{equation}
\mathcal{L}_{\text{char}}
=
\sum_{\mathbf{p}}
\rho\!\left(
\boldsymbol\omega(\mathbf{p}) \odot
\left(\hat{\textbf{I}}_{t}(\mathbf{p})-\textbf{I}_{t}^{gt}(\mathbf{p})\right)
\right), \mbox{ and  }
\rho(x)=\left(x^2+\epsilon^2\right)^{\alpha},
\end{equation}
where $\alpha=0.5$, $\epsilon=10^{-3}$.
The weight function is defined as:
\begin{equation}
\boldsymbol\omega(\mathbf{p})=\cos(\boldsymbol\theta_{\mathbf{p}}), \qquad
\boldsymbol\theta_{\mathbf{p}}=\pi\left(\frac{v_{\mathbf{p}}}{H}-\frac{1}{2}\right),
\end{equation}
where $v_{\mathbf{p}}$ is the vertical pixel coordinate of $\mathbf{p}$ and $H$ is image
height. This weighting reduces the dominance of heavily distorted polar regions while
preserving stronger supervision near the equator. Please see the Supplementary Material for the details of the impact of this loss design and differences to AMT.
\section{Experiments}
\label{sec:experiments}
\subsection{Benchmarks and Evaluation Metrics}
We evaluate SVI360 on two synthetic omnidirectional datasets (FlowScape~\cite{flowscape} and Flow360~\cite{flow360}) and two real-world datasets (ODV360~\cite{odv360} and 360VFI~\cite{360vfi}). The 
 synthetic datasets provide dense ground-truth optical flow, making them well suited for both qualitative and
  quantitative evaluation of optical flow accuracy and frame interpolation quality. ODV360 and 360VFI consist of real-world 360$^\circ$ videos captured in unconstrained environments with complex object motion. They serve as challenging benchmarks for assessing the robustness of interpolation methods in realistic scenarios.

\begin{itemize}
    
\item \textbf{FlowScape}: middle-frame interpolation benchmark with 4,900 training and 1,372 testing triplets.  

\item \textbf{Flow360}: arbitrary-timestep interpolation benchmark with 9-frame tuples (1,286 training, 712 testing).  

\item \textbf{ODV360}: real-world triplets split into 18,344/1,894/1,894 for train/val/test.  

\item \textbf{360VFI}: real-world benchmark with 10,074 training and 930 testing triplets, with test subsets Easy/Middle/Hard/Extreme by motion magnitude.
\end{itemize}
We measure interpolation quality using standard perspective-image metrics, 
including PSNR and SSIM, and assess optical-flow accuracy with EPE and AE.
We also consider spherical-aware metrics: Spherical end-point error (SEPE\cite{priorflow}) 
and weighted-to-spherically-uniform image-quality metrics, namely WS-PSNR \cite{7961186}
and WS-SSIM \cite{8652269}. These weighted quality metrics account for the non-uniform spherical distortion, allowing a fairer assessment of each model’s true performance by reducing the bias from larger errors near the poles.
 For optical flow, SEPE is defined as the average geodesic distance between predicted and ground-truth end points of flow vectors on the unit sphere.

\subsection{Implementation Details}
All experiments were conducted using two H100 GPUs (80\,GB). 
Models are optimized with AdamW with a cosine learning-rate schedule that decays from 
$2 \times 10^{-4}$ to $2 \times 10^{-5}$ over 300 epochs with a batch size of 16. 
The loss weight $\lambda$ is fixed to $1$.
Data augmentation is done with horizontal and vertical flips, color jitter, and temporal reversal (swapping $I_0$/$I_1$), and a random erasing module that masks regions in the third frame to mimic occlusions caused by optical-flow warping. Besides that, we also apply spherical-specific augmentation method: random roll, pitch, yaw rotations, to create diverse view angles while preserving scene content. These augmentation methods better expose the model to occluded regions, wrap-around discontinuities, and viewpoint variations.

\subsection{Comparison and Results}
We compare SVI360 against SuperSloMo~\cite{superslomo}, FILM~\cite{film}, RIFE~\cite{rife}, VFIFormer~\cite{vfiformer}, IFRNet-L~\cite{ifrnet}, and AMT-G~\cite{amt}. For fair comparison, these methods are retrained on our main benchmarks (FlowScape, Flow360, ODV360). We report PSNR, SSIM, WS-PSNR, and WS-SSIM on all datasets, while optical flow metrics (EPE, AE, and SEPE) are reported only on FlowScape. We do not retrain 360VFI~\cite{360vfi} because reproducible training code is unavailable. On the 360VFI dataset we train our method and the strong baseline AMT-G, then compare against published baselines, including 360VFI~\cite{360vfi}, DQBC~\cite{ijcai2023p198}, EMA-VFI~\cite{zhang2023extracting}, EBME~\cite{Jin_2023_WACV}, UPR-Net~\cite{Jin_2023_CVPR}, and IFRNet~\cite{ifrnet}, under the official benchmark split. \vspace{0.2cm}

\noindent\textbf{FlowScape -- Middle timestep interpolation ($t=0.5$)}. \\
In this setting, we report both interpolation quality and optical flow accuracy in ~\cref{tab:flowscape_results_main}.
\begin{table*}[t]
  \centering
  \caption{Results on FlowScape. Higher is better for PSNR/WS-PSNR/SSIM/WS-SSIM; lower is better for EPE/SEPE/AE. For each group, the best result is shown in \textbf{bold}, and the second best is \underline{underlined}.}
  \label{tab:flowscape_results_main}
  \resizebox{\textwidth}{!}{
  \begin{tabular}{lccccccccc}
    \toprule
    Method &
    PSNR &
    WS-PSNR &
    SSIM &
    WS-SSIM &
    EPE &
    SEPE &
    AE &
    \shortstack{Params\\(M)} &
    \shortstack{Latency\\(ms/f)} \\
    \midrule
    SuperSloMo & 33.11 & 32.48 & 0.9034 & 0.9045 & 15.79 & 23.74 & 29.76 & 19.8 & 62 \\
    FILM & 35.98 & 35.84 & 0.9369 & 0.9282 & 14.65 & 20.26 & 23.19 & 34.4 & 393 \\
    RIFE & 35.97 & 35.66 & 0.9540 & 0.9438 & 15.16 & 21.70 & 23.56 & 9.8 & 26 \\
    VFIFormer & 36.90 & 36.78 & 0.9614 & 0.9546 & 14.07 & 20.12 & 22.84 & 24.1 & 1293 \\
    IFRNet-L & 37.03 & 36.58 & 0.9651 & 0.9564 & 12.64 & \underline{15.86} & \underline{20.23} & 19.7 & 79 \\
    AMT-G & \underline{38.05} & \underline{37.40} & \underline{0.9704} & \underline{0.9622} & \textbf{12.26} & 25.91 & 20.90 & 30.6 & 250 \\
    \textbf{SVI360} (Ours) & \textbf{38.68} & \textbf{37.73} & \textbf{0.9754} & \textbf{0.9647} & \underline{12.36} & \textbf{15.22} & \textbf{19.74} & 50.1 & 365 \\
    \bottomrule
  \end{tabular}
  }
\end{table*}
Overall, SVI360 consistently outperforms the competitors across most evaluated
metrics. AMT-G remains slightly better on EPE, but SVI360 provides a stronger overall
balance between interpolation fidelity and spherical motion consistency.
Compared with the second-best method, SVI360 surpasses it by 0.63\,dB in PSNR and
0.33\,dB in WS-PSNR. Beyond the global gains, the improvements in WS-PSNR and WS-SSIM indicate better
perceptual quality over the full panoramic frame under distortion-aware weighting,
rather than gains limited to specific regions. AMT-G remains a strong competitor, but it
uses explicit flow supervision, whereas SVI360 does not rely on this auxiliary
supervision. Earlier baselines show larger degradation on panoramic data. Due to its
dual-branch design, SVI360 uses more parameters than most baselines, yet it still maintains
a reasonable latency, yielding a competitive accuracy-efficiency tradeoff.
\begin{figure*}[t]
  \centering
  \scriptsize
  \setlength{\tabcolsep}{2pt}
  \renewcommand{\arraystretch}{1.0}

  \begin{tabular}{c c c c}
    \textbf{GT} & \textbf{SVI360} (Ours) & \textbf{AMT-G} & \textbf{IFRNet-L} \\
    \includegraphics[width=0.23\textwidth]{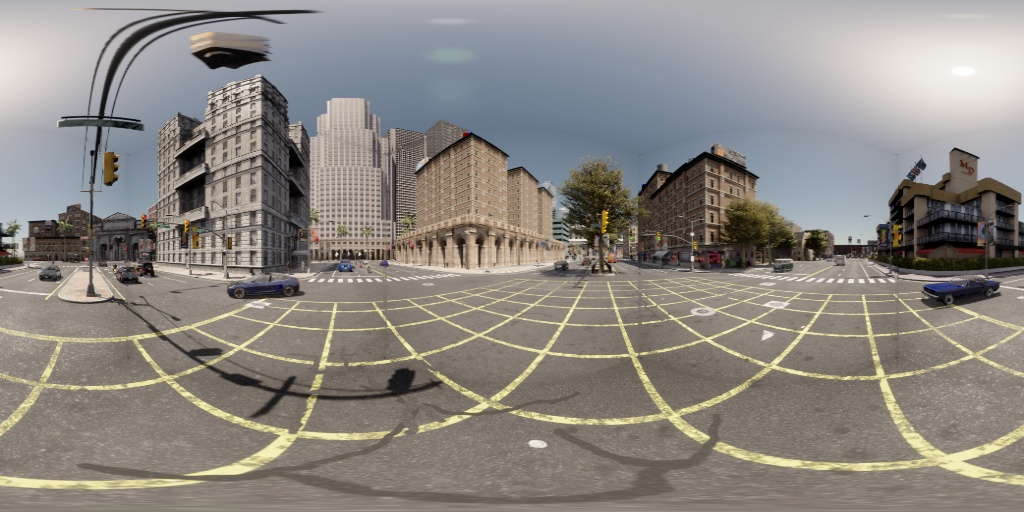} &
    \includegraphics[width=0.23\textwidth]{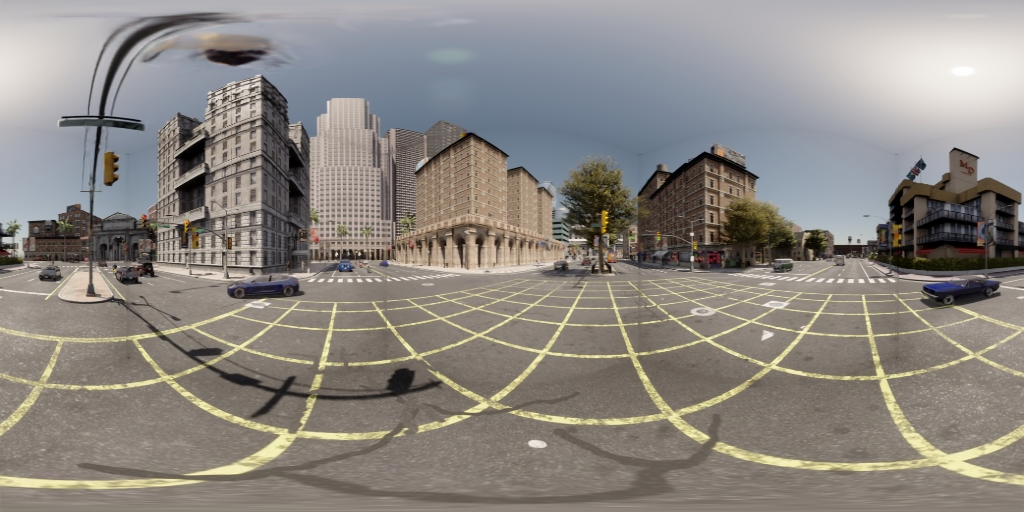} &
    \includegraphics[width=0.23\textwidth]{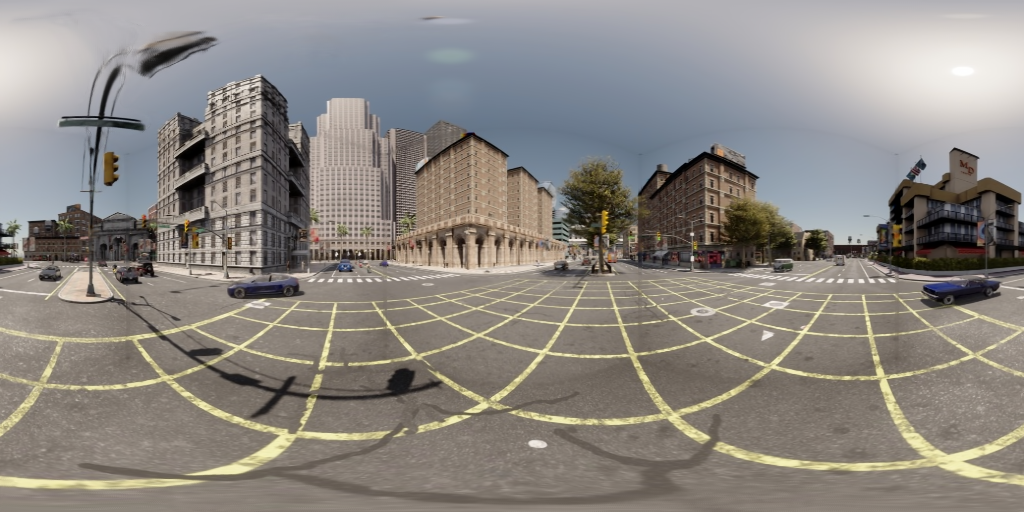} &
    \includegraphics[width=0.23\textwidth]{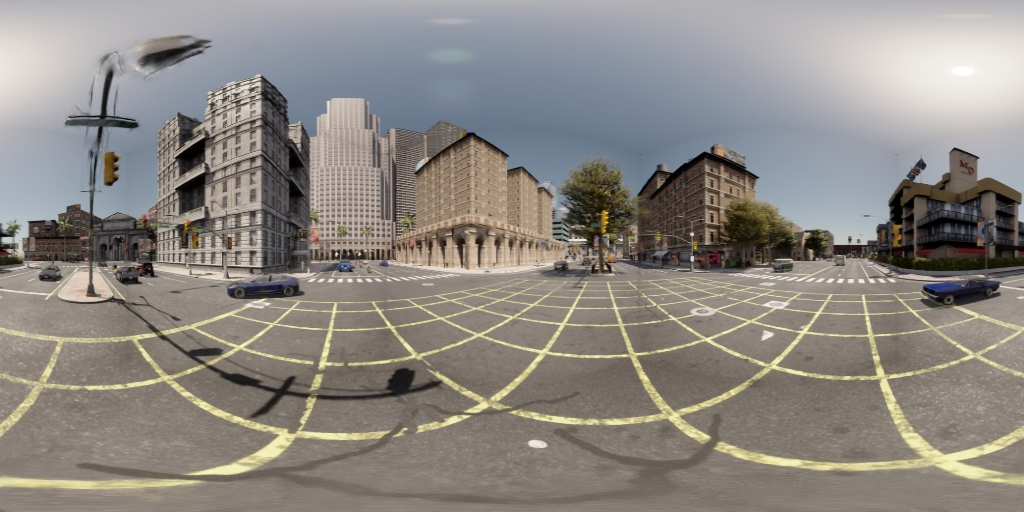} \\
    {\tiny --} & {\tiny PSNR 29.88, SSIM 0.9593} & {\tiny PSNR 25.73, SSIM 0.9441} & {\tiny PSNR 25.51, SSIM 0.9322} \\
    \includegraphics[width=0.23\textwidth]{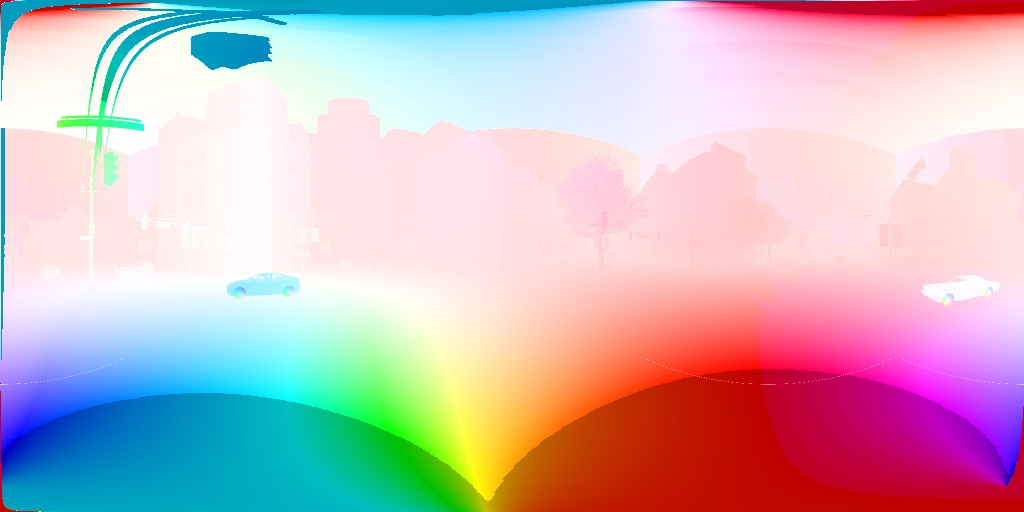} &
    \includegraphics[width=0.23\textwidth]{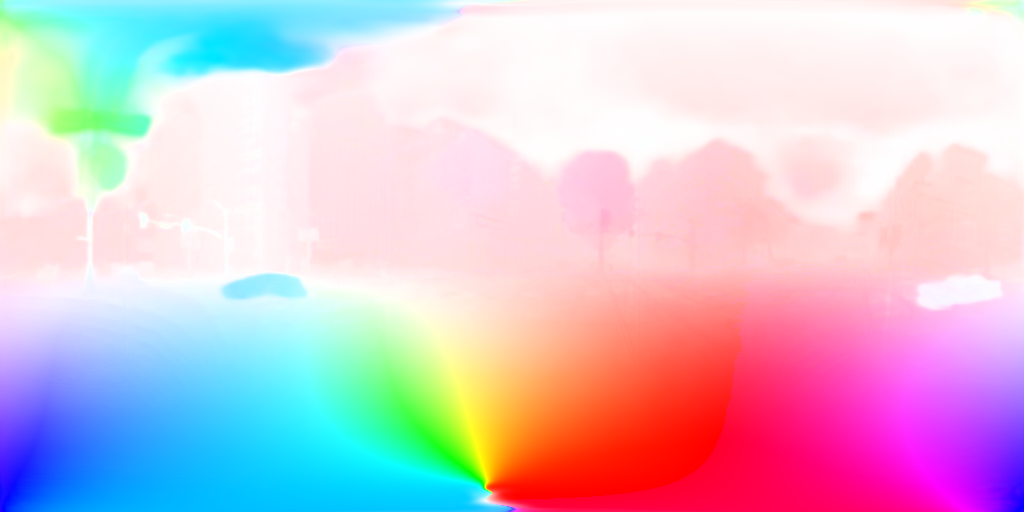} &
    \includegraphics[width=0.23\textwidth]{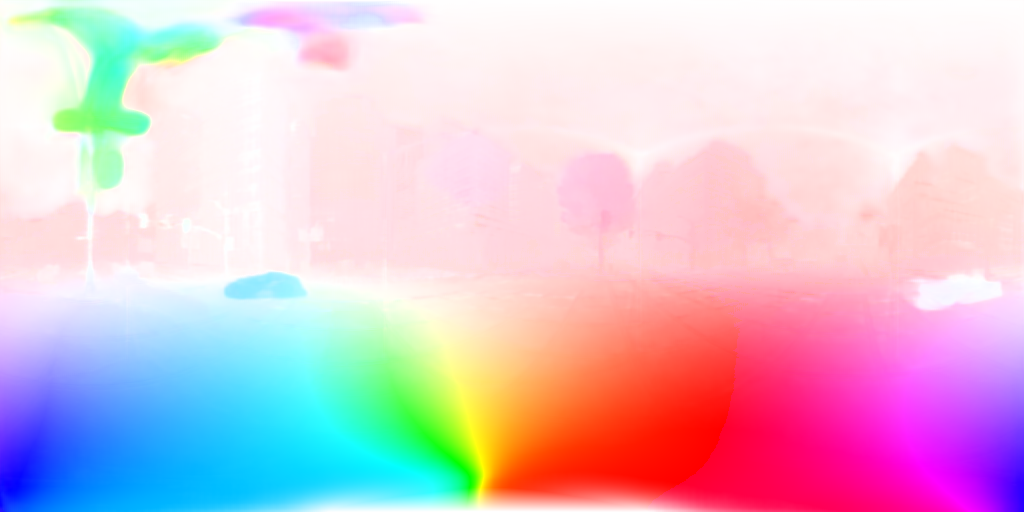} &
    \includegraphics[width=0.23\textwidth]{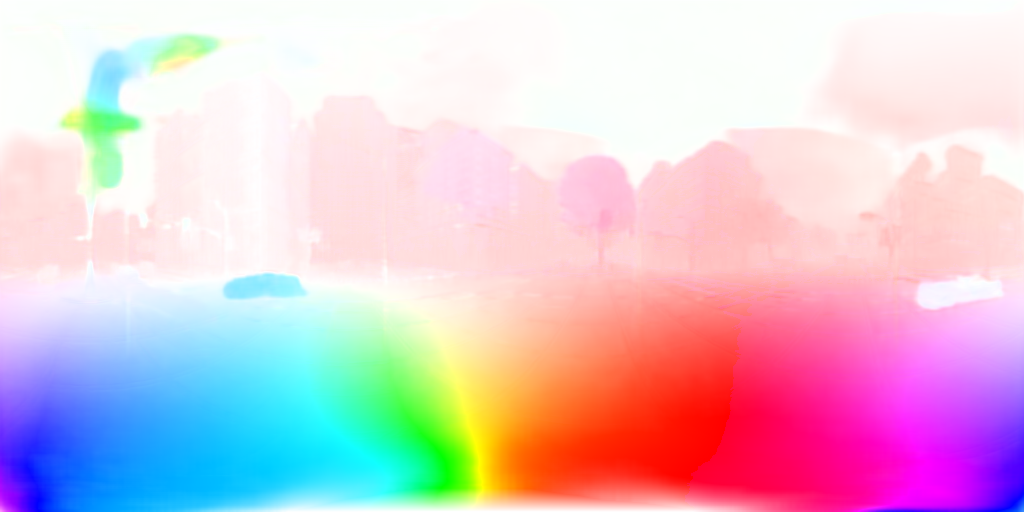} \\
    {\tiny --} & {\tiny EPE 17.24, AE 22.75} & {\tiny EPE 16.33, AE 24.67} & {\tiny EPE 16.85, AE 23.70} \\
  \end{tabular}

  \vspace{2mm}

  \begin{tabular}{c c c c}
    \textbf{VFIFormer} & \textbf{RIFE} & \textbf{FILM} & \textbf{SuperSloMo} \\
    \includegraphics[width=0.23\textwidth]{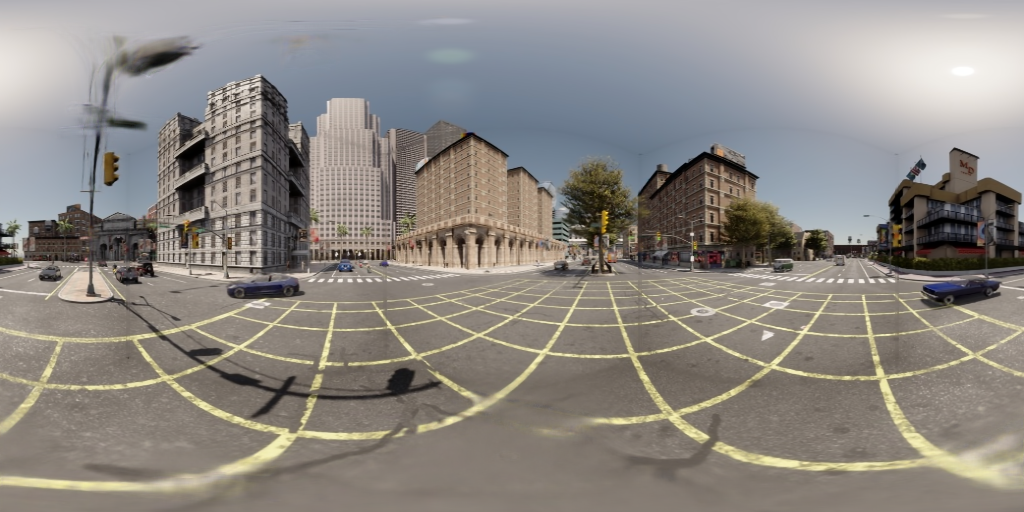} &
    \includegraphics[width=0.23\textwidth]{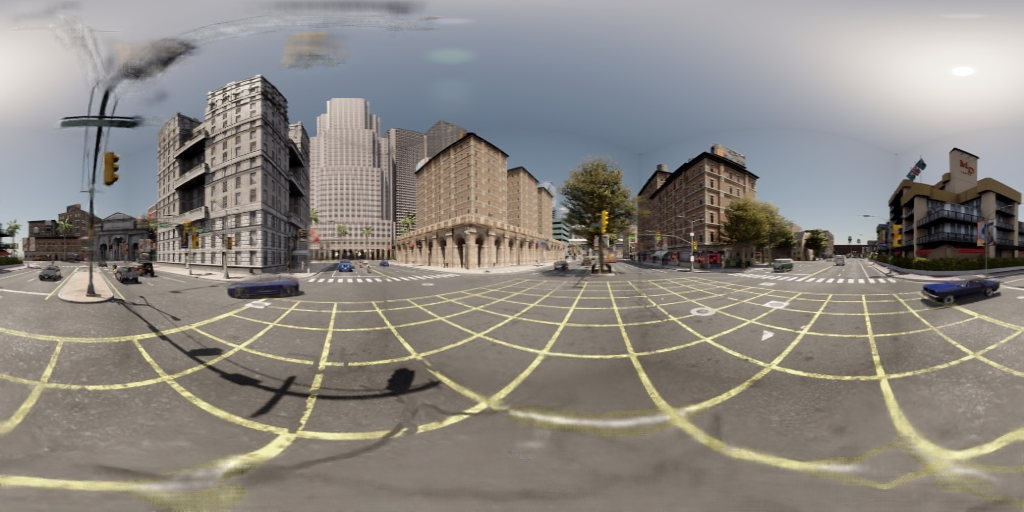} &
    \includegraphics[width=0.23\textwidth]{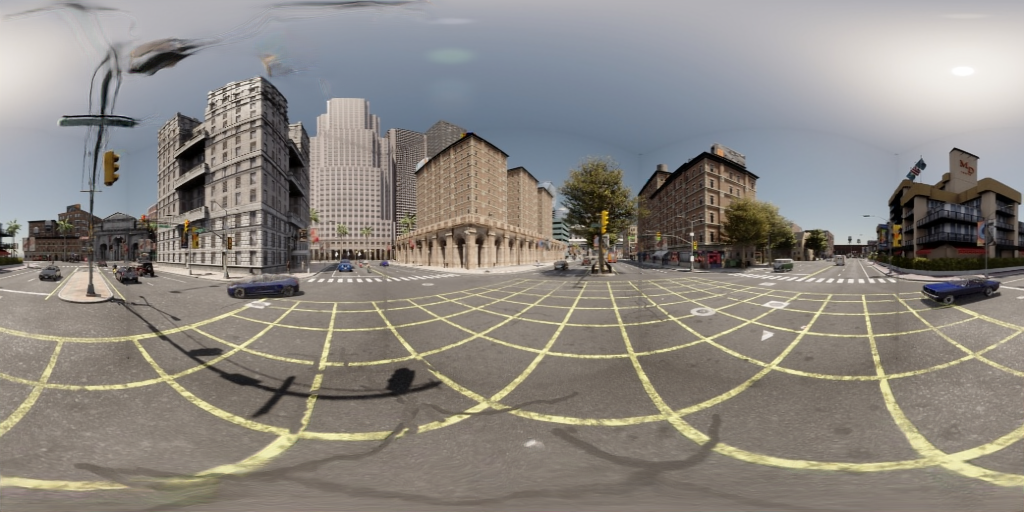} &
    \includegraphics[width=0.23\textwidth]{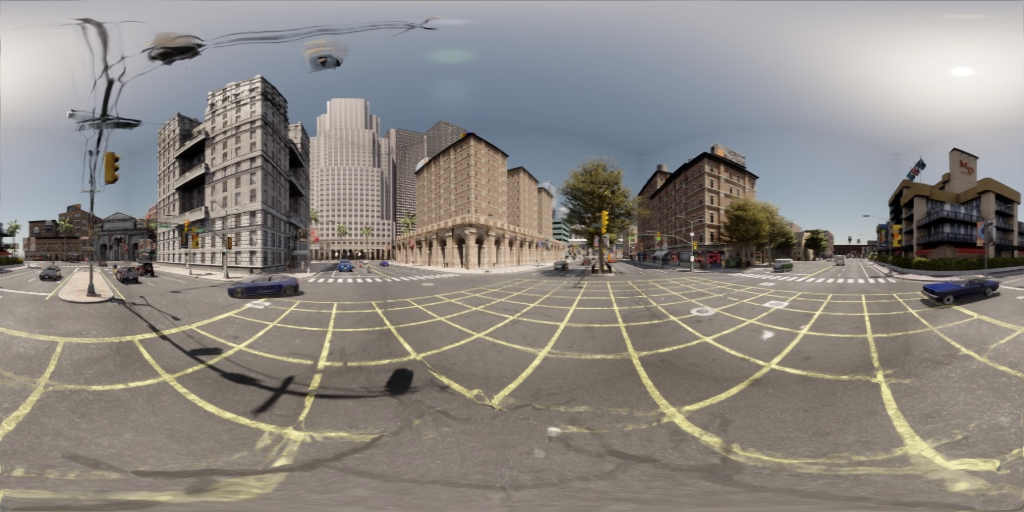} \\
    {\tiny PSNR 24.63, SSIM 0.9142} & {\tiny PSNR 24.39, SSIM 0.8906} & {\tiny PSNR 24.77, SSIM 0.8697} & {\tiny PSNR 22.45, SSIM 0.7769} \\
    \includegraphics[width=0.23\textwidth]{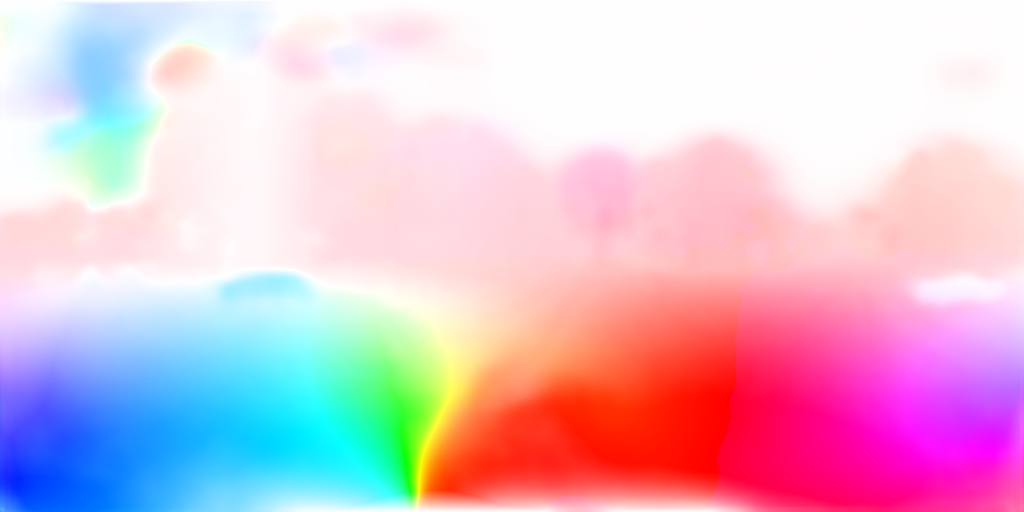} &
    \includegraphics[width=0.23\textwidth]{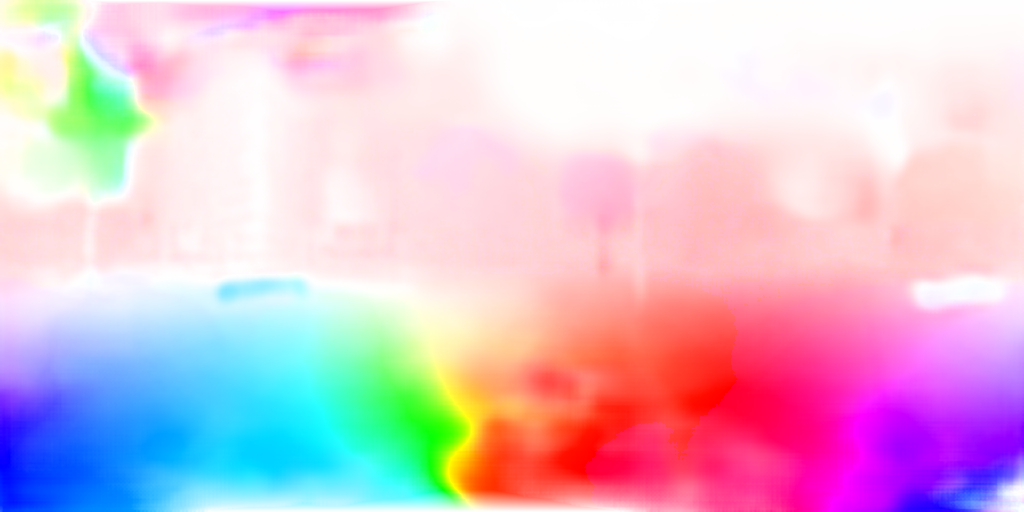} &
    \includegraphics[width=0.23\textwidth]{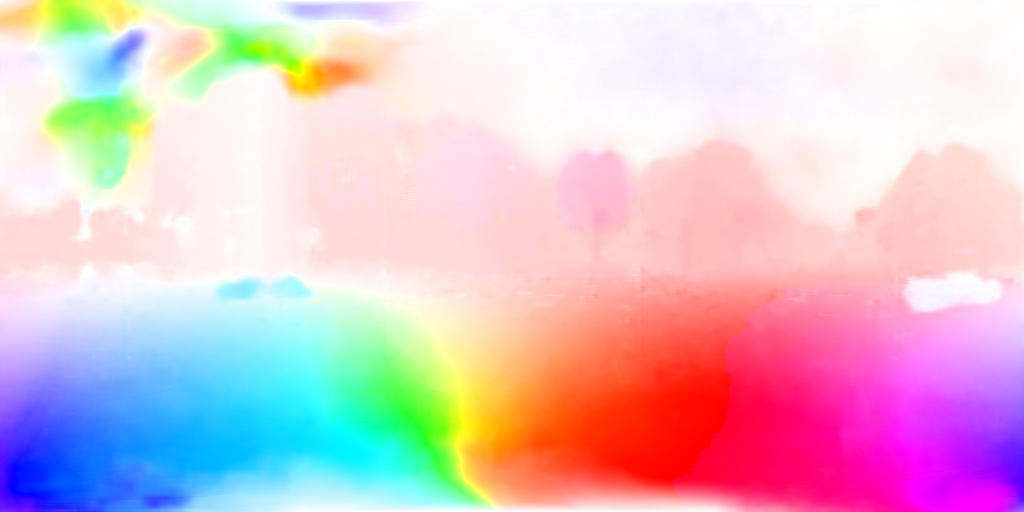} &
    \includegraphics[width=0.23\textwidth]{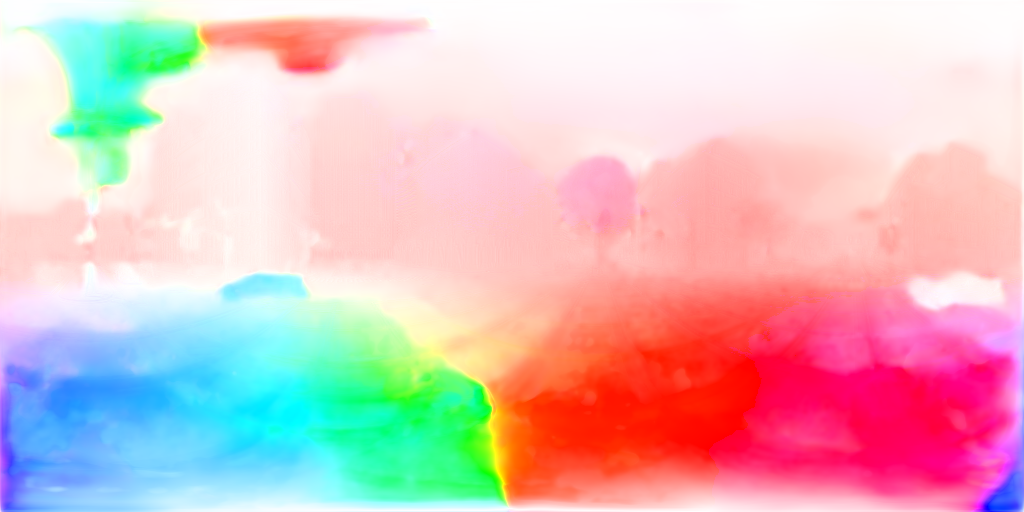} \\
    {\tiny EPE 18.62, AE 30.38} & {\tiny EPE 19.99, AE 29.85} & {\tiny EPE 18.82, AE 25.83} & {\tiny EPE 22.08, AE 31.28} \\
  \end{tabular}

  \caption{Qualitative comparison on FlowScape. For each method, the upper image is the interpolated middle frame and the lower image is the estimated optical flow. The values shown below images report PSNR/SSIM for interpolation and EPE/AE for the flow of each image sample (not dataset-level metrics).}
  \label{fig:flowscape_qualitative}
\end{figure*}

As shown in Fig.~\ref{fig:flowscape_qualitative}, SVI360 also preserves finer structural details and sharper motion boundaries, with fewer ghosting and blending artifacts than competing methods, especially
near highly distorted polar regions. Benefiting from multi-view priors, where
complementary views refine each other, SVI360 better maintains fine object structures at
the poles while keeping high-quality reconstruction in other regions. For optical flow, SVI360 performs on par with strong baselines such as AMT-G and IFRNet-L, while other methods show more severe degradation in difficult areas with large pixel displacement and occlusion. It is also worth noting that estimated motion does not need to be maximally similar to the ground-truth flow at every local detail. As discussed in \cite{Xue_2019}, highly precise optical flow estimation can be intractable and even
suboptimal for a specific downstream video task. Similar to the design philosophy of
AMT \cite{amt} and IFRNet \cite{ifrnet}, our objective is to estimate flow that is globally
consistent with ground truth while allowing task-beneficial local diversity. The
predicted flow of SVI360 satisfies this objective. Overall, the qualitative and quantitative comparison
suggests that SVI360 improves both visual reconstruction fidelity and motion consistency
on FlowScape.

\noindent\textbf{Flow360 -- Arbitrary timestep interpolation}. \\
Using 9-frame tuples, target timestamps are $t\in\{1/8,2/8,\ldots,7/8\}$. Methods marked with (*) can only predict the
center frame at $t=0.5$. For these methods, we use recursive inference to obtain 8$\times$
interpolation: first predict $I_{0.5}$ from $(I_0,I_1)$, then predict $I_{0.25}$ from
$(I_0,I_{0.5})$ and $I_{0.75}$ from $(I_{0.5},I_1)$, and continue similarly to obtain the
remaining intermediate frames.
On Flow360 (Table~\ref{tab:flow360_odv360_results_main}), SVI360 achieves the best
performance across all interpolation metrics, showing strong robustness for
arbitrary-timestep prediction. It maintains at least a 0.25\,dB PSNR gain against strong competitors. FILM and VFIformer are additionally affected by recursive inference for non-center timestamps,
which can accumulate errors and reduce quality. Overall, these results confirm
that SVI360 is very effective for multiple frame interpolation.
\begin{table*}[t]
  \centering
  \caption{Results on Flow360 and ODV360. Higher is better for PSNR/SSIM/WS-PSNR/WS-SSIM. For each dataset, the best result is shown in \textbf{bold}, and the second best is \underline{underlined}. (*) means the methods can only predict the center frame.}
  \label{tab:flow360_odv360_results_main}
  \begin{tabular}{lcccc|cccc}
    \toprule
    & \multicolumn{4}{c}{Flow360} & \multicolumn{4}{c}{ODV360} \\
    \cmidrule(lr){2-5} \cmidrule(lr){6-9}
    Method & PSNR & WS-PSNR & SSIM & WS-SSIM & PSNR & WS-PSNR & SSIM & WS-SSIM \\
    \midrule
    SuperSloMo & 31.43 & 31.24 & 0.9481 & 0.9440 & 26.41 & 26.95 & 0.8816 & 0.8689 \\
    FILM (*) & 32.73 & 32.14 & 0.9514 & 0.9451 & 27.16 & 27.81 & 0.9059 & 0.8703 \\
    RIFE & 32.83 & 32.23 & 0.9631 & 0.9515 & 28.74 & 29.51 & 0.9243 & 0.8992 \\
    VFIFormer (*) & 32.54 & 31.93 & 0.9614 & 0.9502 & 27.85 & 28.36 & 0.9116 & 0.8725 \\
    IFRNet-L & 33.16 & 32.60 & 0.9656 & 0.9529 & 28.18 & 28.69 & 0.9167 & 0.8830 \\
    AMT-G & \underline{33.77} & \underline{33.22} & \underline{0.9708} & \underline{0.9594} & \underline{29.07} & \underline{29.75} & \underline{0.9303} & \underline{0.9067} \\
    Ours & \textbf{34.02} & \textbf{33.30} & \textbf{0.9752} & \textbf{0.9615} & \textbf{29.25} & \textbf{29.93} & \textbf{0.9332} & \textbf{0.9105} \\
    \bottomrule
  \end{tabular}
\end{table*}

\vspace{0.2cm}
\noindent\textbf{ODV360 -- Middle frame interpolation in real-world scenes}. \\
The model must remain robust to real capture artifacts. On the real-world ODV360 benchmark (Table~\ref{tab:flow360_odv360_results_main}), SVI360
also achieves the best results, indicating strong generalization to
unconstrained panoramic scenes with illumination variation and complex non-rigid motion.
\begin{figure*}[t]
  \centering
  \scriptsize
  \setlength{\tabcolsep}{2pt}
  \renewcommand{\arraystretch}{1.0}
  \newcommand{\odvtopbox}[1]{%
    \begin{tikzpicture}[baseline=(img.base)]
      \node[inner sep=0] (img) {\includegraphics[width=0.32\textwidth]{#1}};
      \draw[red, line width=1.1pt]
        ([xshift=0.1cm,yshift=-0.01cm]img.north west)
        rectangle
        ([xshift=-0.1cm,yshift=-0.35cm]img.north east);
    \end{tikzpicture}%
  }
  \begin{tabular}{ccc}
    \multicolumn{3}{c}{ODV360 representative sample} \\
    \begin{tabular}{c}
      \textbf{Inputs (Overlay)} \\
      \odvtopbox{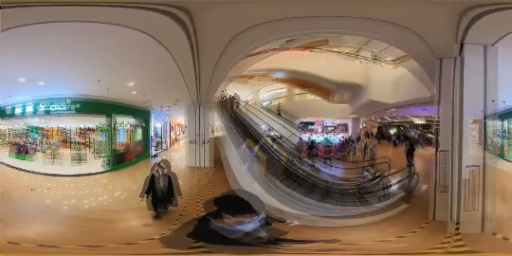} \\
      {\tiny --}
    \end{tabular} &
    \begin{tabular}{c}
      \textbf{SVI360 (ours)} \\
      \odvtopbox{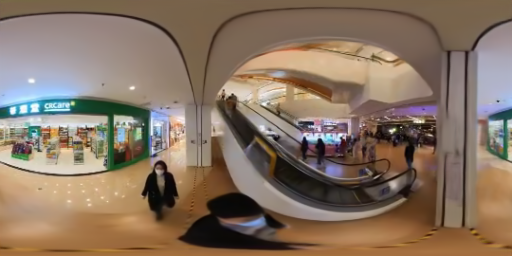} \\
      {\tiny PSNR 26.58, SSIM 0.9292}
    \end{tabular} &
    \begin{tabular}{c}
      \textbf{AMT-G} \\
      \odvtopbox{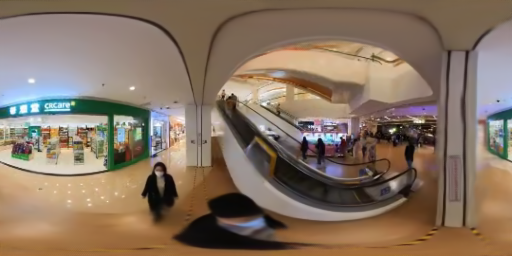} \\
      {\tiny PSNR 26.01, SSIM 0.9207}
    \end{tabular} \\
    \begin{tabular}{c}
      \textbf{IFRNet-L} \\
      \odvtopbox{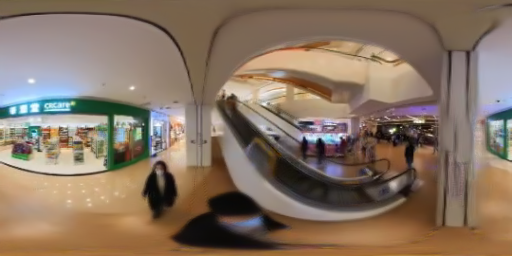} \\
      {\tiny PSNR 25.02, SSIM 0.8938}
    \end{tabular} &
    \begin{tabular}{c}
      \textbf{RIFE} \\
      \odvtopbox{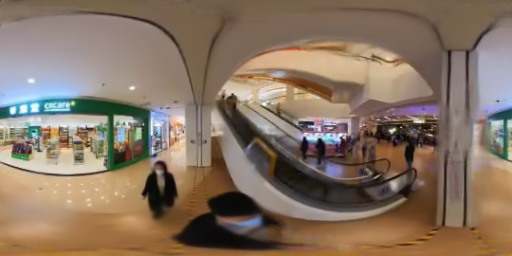} \\
      {\tiny PSNR 25.05, SSIM 0.8993}
    \end{tabular} &
    \begin{tabular}{c}
      \textbf{VFIFormer} \\
      \odvtopbox{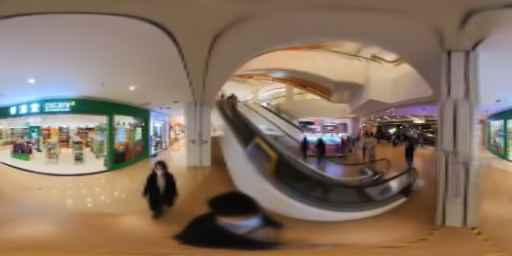} \\
      {\tiny PSNR 24.81, SSIM 0.8881}
    \end{tabular} \\
  \end{tabular}

  \caption{Qualitative comparison on ODV360. The highlighted top region marks a challenging distorted area.}
  \label{fig:odv360_qualitative}
\end{figure*}
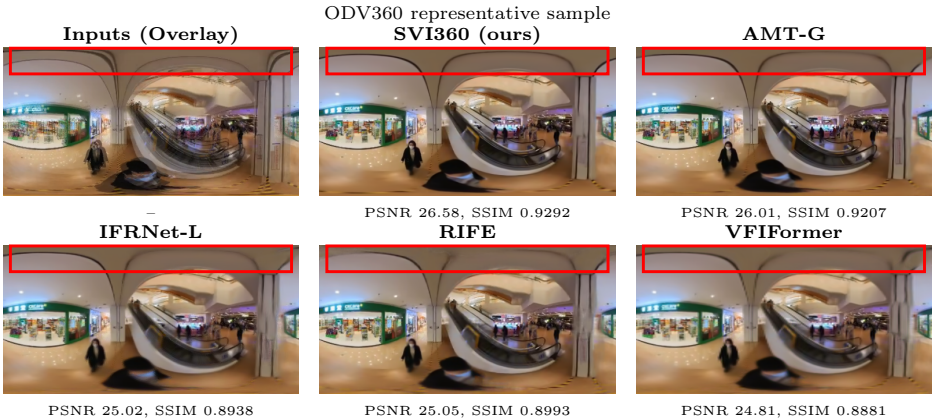
As can be seen in Fig.~\ref{fig:odv360_qualitative}, SVI360 yields the highest per-image PSNR/SSIM and
better preserves the ceiling structure in the top region, which is consistent with the GT image.

\vspace{0.2cm} \noindent\textbf{360VFI -- Interpolation across varying difficulty levels}. \\
We train our method and strong baseline AMT-G on the 360VFI training set and compare it with published results reported in \cite{360vfi} and other prior methods listed in Table~\ref{tab:comparison_360vfi_benchmark}.
\begin{table*}[!t]
  \centering
  \caption{Quantitative comparison on the 360VFI benchmark across increasing difficulty levels. Higher is better for both metrics.
  }
  \label{tab:comparison_360vfi_benchmark}
  \resizebox{\textwidth}{!}{
  \begin{tabular}{lcccc}
    \toprule
    Method & Easy & Middle & Hard & Extreme \\
    \cmidrule(lr){2-5}
    & WS-PSNR/WS-SSIM & WS-PSNR/WS-SSIM & WS-PSNR/WS-SSIM & WS-PSNR/WS-SSIM \\
    \midrule
    IFRNet~\cite{ifrnet} & 33.90/\underline{0.9538} & 28.89/0.9047 & 27.74/0.8724 & 24.80/0.7999 \\
    DQBC~\cite{ijcai2023p198} & 33.51/0.9449 & 27.35/0.8776 & 26.02/0.8329 & 23.26/0.7410 \\
    EMA-VFI~\cite{zhang2023extracting} & 33.40/0.9529 & 28.37/0.9024 & 27.64/0.8817 & 24.89/0.8161 \\
    EBME~\cite{Jin_2023_WACV} & 33.48/0.9447 & 27.35/0.8772 & 25.97/0.8318 & 23.23/0.7403 \\
    UPR-Net~\cite{Jin_2023_CVPR} & 33.45/0.9535 & 28.42/0.9024 & 27.72/0.8816 & 24.85/0.8157 \\
    360VFI~\cite{360vfi} & 33.95/0.9537 & 28.96/0.9060 & 27.81/0.8879 & 25.63/0.8517 \\
    AMT-G~\cite{amt} & \textbf{34.84}/0.9726 & 29.74/0.9376 & 28.58/0.9209 & 25.55/0.8817 \\
    Ours & 34.81/\textbf{0.9731} & \textbf{29.98}/\textbf{0.9409} & \textbf{28.95}/\textbf{0.9264} & \textbf{25.92}/\textbf{0.8880} \\
    \bottomrule
  \end{tabular}
  }
\end{table*}
\new{We observe that SVI360 achieves the best performance across all metrics under different difficulty settings. In the Easy setting (flow magnitude < 2), where pixel displacement is minimal, most methods achieve strong performance. However, starting from the Middle setting, as the flow magnitude increases, the performance of all methods declines noticeably. Although our method also experiences some degradation, the drop-off is less significant than that of 360VFI and AMT-G, demonstrating stronger robustness to large motion fields. Qualitative results comparing SVI360 and AMT-G on the 360VFI benchmark are shown in \cref{fig:qualitative_360vfi}. Since the official implementation of 360VFI is not yet publicly available, we could not generate its corresponding qualitative results for comparison.}

\newcommand{\vfiimg}[1]{\includegraphics[width=0.29\textwidth]{#1}}
\begin{figure*}[tb]
\centering
\begin{tabular}{cccc}
 & \textbf{GT} & \textbf{SVI360} & \textbf{AMT-G} \\

Easy &
\vfiimg{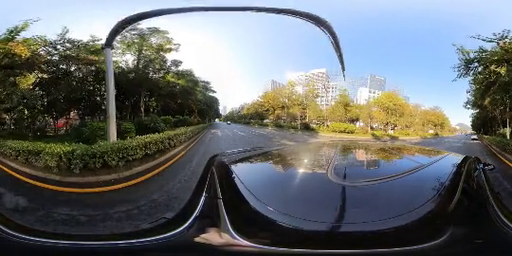} &
\vfiimg{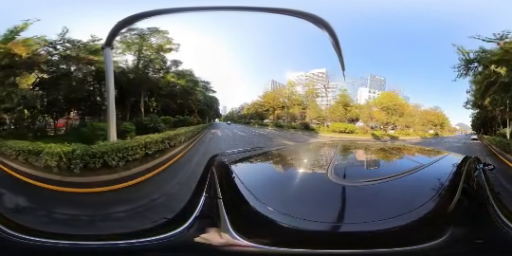} &
\vfiimg{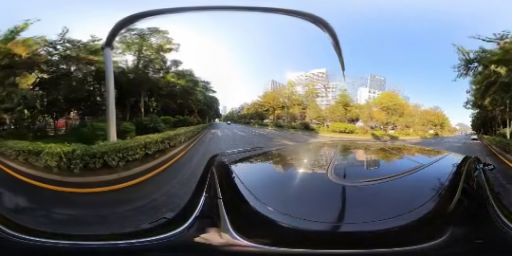} \\

Medium &
\vfiimg{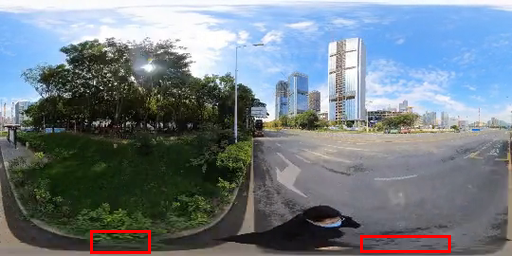} &
\vfiimg{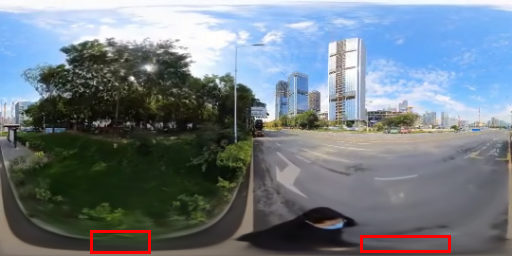} &
\vfiimg{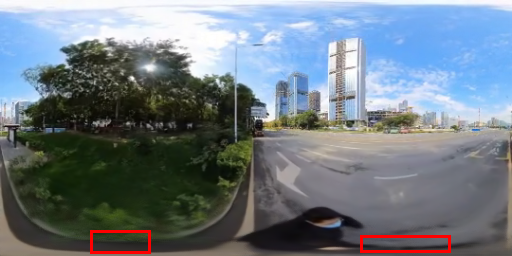} \\

Hard &
\vfiimg{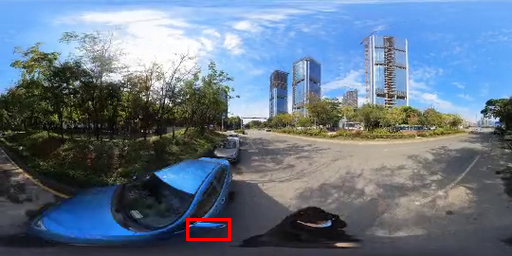} &
\vfiimg{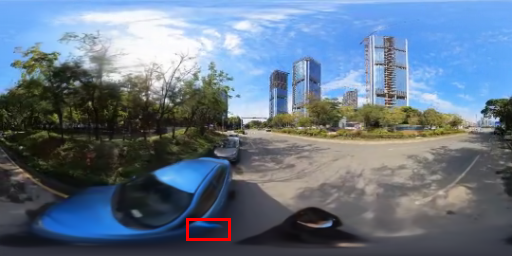} &
\vfiimg{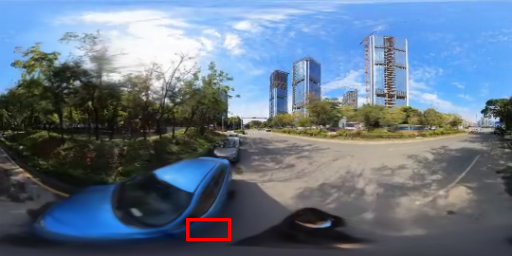} \\

Extreme &
\vfiimg{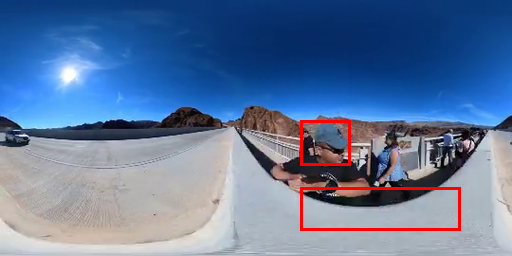} &
\vfiimg{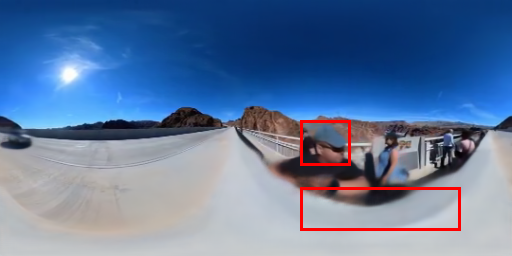} &
\vfiimg{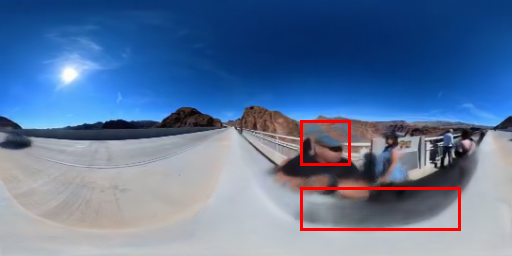}
\end{tabular}

\caption{Qualitative results on the 360VFI benchmark. Red bounding boxes highlight regions where AMT-G fails while SVI360 produces more accurate results.
}
\label{fig:qualitative_360vfi}
\end{figure*}
\subsection{Ablation Study}
All ablation models are trained and evaluated on FlowScape dataset. 
\paragraph{1) Ablation on the Loss Functions.}
We evaluate the contribution of the spherical-weighted Charbonnier and the census term. This analysis quantifies how each loss element affects interpolation quality on spherical data.
\paragraph{2) Ablation on SVI360 Architecture.}
We analyze the impact of key architectural modules, including the Spherical Refiner
(SR) block and the Spherical Multi-Field Refinement module. For the latter, instead of using two branches as in \cref{eq:fuse}, we remove orthogonal-branch
candidates and fuse only primitive-branch candidates, i.e.,
$\textbf{I}_t=\mathcal{F}_{\text{fuse}}\!\left(\{\textbf{I}_{t}^{p,1},\ldots,\textbf{I}_{t}^{p,N}\}\right)$.
This setting isolates the contribution of orthogonal-branch multi-field cues. Together
with the SR-block removal, it shows the effect of each component on overall
interpolation performance.

\begin{figure}[t]
  \centering
  \begin{minipage}[t]{0.18\linewidth}
    \centering
    \includegraphics[width=0.90\linewidth,trim=110 120 720 150,clip]{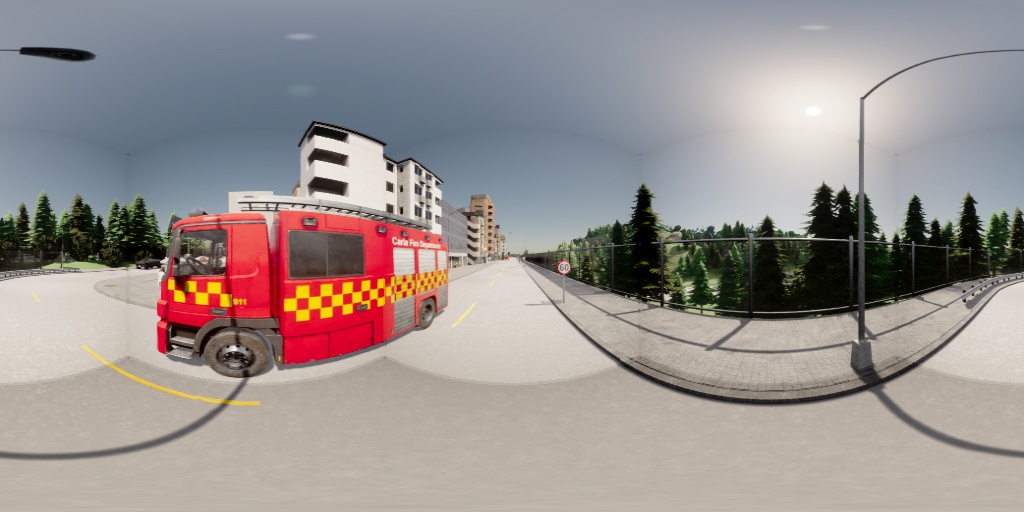}
    \vspace{0.5mm}
    {\scriptsize GT}
  \end{minipage}
  \hspace{0.015\linewidth}
  \begin{minipage}[t]{0.18\linewidth}
    \centering
    \includegraphics[width=0.90\linewidth,trim=110 120 720 150,clip]{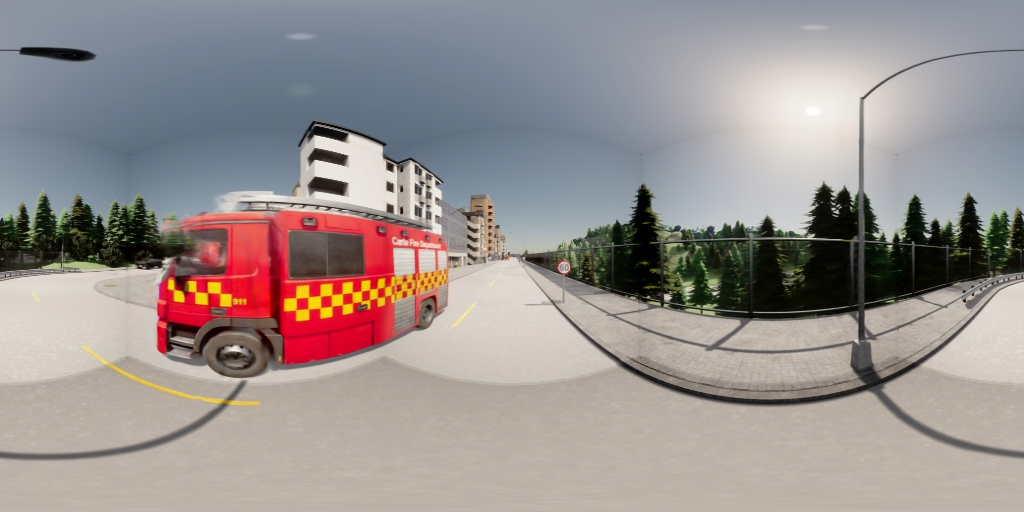}
    \vspace{0.5mm}
    {\scriptsize Standard loss}
  \end{minipage}
  \hspace{0.015\linewidth}
  \begin{minipage}[t]{0.18\linewidth}
    \centering
    \includegraphics[width=0.90\linewidth,trim=110 120 720 150,clip]{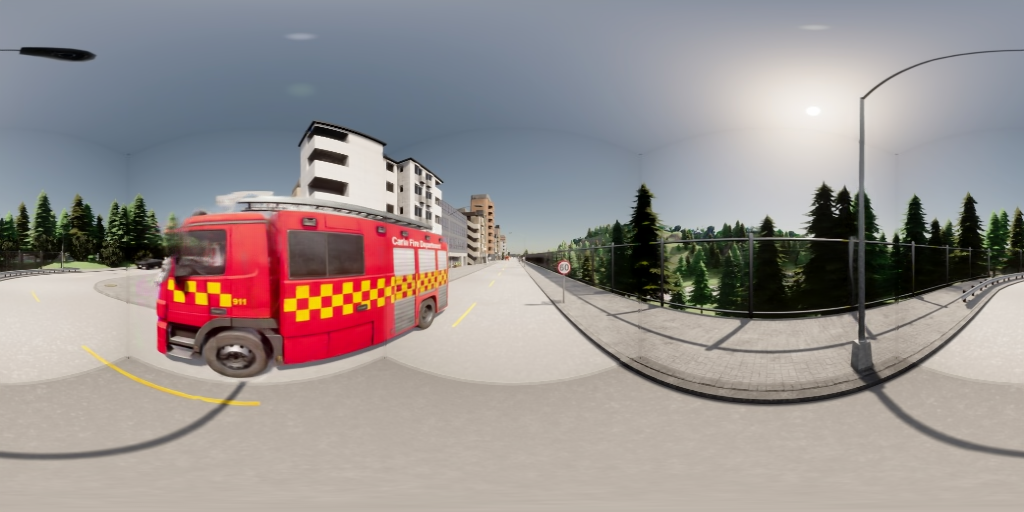}
    \vspace{0.5mm}
    {\tiny Spherical Loss}
  \end{minipage}\vspace{-0.2cm}
  \caption{Qualitative comparison of reconstruction losses on FlowScape. The
  images show a center-region zoom of the ground truth (left) and reconstructions produced
  with standard Charbonnier loss (center) and spherical weighted Charbonnier loss (right). Please notice the spherical Charbonnier loss better preserves the firetruck cabin structure.}
  \label{fig:ablation_spherical_loss}\vspace{-0.3cm}
\end{figure}

\begin{table*}[!tb]
  \centering
  \caption{Ablation on loss functions and SVI360 architecture. Higher is better.}
  \label{tab:ablation_merged}
  \begin{tabular}{lcccc}
    \toprule
    Settings & PSNR & WS-PSNR & SSIM & WS-SSIM \\
    \midrule \multicolumn{5}{l}{\textbf{Loss Functions}} \\ w/o sph. Charb. Loss & 33.99 & 33.85 & 0.9099 & 0.9471 \\ w/o Census Loss & 38.34 & 37.63 & 0.9734 & 0.9644 \\ Full loss & \textbf{38.68} & \textbf{37.73} & \textbf{0.9754} & \textbf{0.9647} \\ \midrule \multicolumn{5}{l}{\textbf{SVI360 Architecture}} \\ w/o sph. multi-field Refinement & 38.08 & 37.29 & 0.9715 & 0.9627 \\ w/o Spherical Refiner (SR) & 38.59 & 37.65 & 0.9748 & 0.9640 \\ Full model & \textbf{38.68} & \textbf{37.73} & \textbf{0.9754} & \textbf{0.9647} \\ \bottomrule
  \end{tabular}
\end{table*}
Table~\ref{tab:ablation_merged} confirms that each design choice contributes to the
final performance. In the loss ablation, removing $\mathcal{L}_{\text{char}}$ causes a clear performance drop, showing that the Charbonnier term is the main reconstruction driver. Without $\mathcal{L}_{\text{cen}}$ term, training is more unstable, therefore, we recommend using both losses together. We can also notice the relevance of the spherical weighted Charbonnier loss to preserve image details of regions near the equator, as shown in \cref{fig:ablation_spherical_loss}.   

For architecture ablation, using only primitive-branch candidates reduces all metrics compared with the full model, which demonstrates the benefit of cross-branch multi-field fusion for spherical interpolation quality. \new{Removing the Spherical Refiner also degrades the interpolation metrics, showing that the aligned orthogonal cues such as estimated flow and intermediate features help improve the synthesized intermediate frame.}
\new{To verify that improvements do not simply come from a larger number of parameters, we also evaluate a lightweight variant, LiteSVI360, in the Supplementary Material. LiteSVI360 uses smaller AMT backbones and has substantially fewer parameters, while maintaining competitive performance, showing that the gains mainly come from the proposed spherical dual-branch refinement and fusion design.}

\vspace{0.2cm}
\noindent\textbf{Discussion and limitations.}
\noindent The dual-branch equivariance feature learning design has shown effectiveness in 360$^\circ$ optical flow
\cite{priorflow}, and our results show that the same principle also transfers to
interpolation. This suggests a broader potential for spherical reconstruction tasks,
including segmentation and depth estimation.
\noindent Limitations: Prior VR findings report that frame rates above 120\,fps can improve user comfort and
reduce simulator sickness \cite{wang2023framrate}. In contrast, our model currently
runs at 365\,ms/frame (nearly 3\,fps), so it is not suitable for real-time VR
interpolation. However, many 360$^\circ$ contents are not rendered online. In these
cases, SVI360 is better aligned with offline pipelines such as post-production video
enhancement and compression-oriented preprocessing where quality is prioritized over
latency.
\section{Conclusion}
\label{sec:conclusion}
This paper presents SVI360, a dual-branch omnidirectional video interpolation method that exploits complementary priors from primitive and
orthogonal spherical views. By combining cross-view refinement with
distortion-aware spherical reconstruction, SVI360 improves both interpolation
fidelity and motion consistency in panoramic scenes. Extensive experiments on
FlowScape, Flow360, ODV360 and 360VFI benchmarks show that SVI360 achieves
strong and consistent performance, with clear gains in interpolation quality
and robust behavior in challenging real-world scenes. High-quality spherical video interpolation
remains fundamentally more challenging than perspective video interpolation, and pushing
visual fidelity in omnidirectional domains currently requires accepting higher
computational cost. Future work will focus on reducing model complexity to meet real-time constraints of VR downstream tasks.\vspace{-0.1cm}

\vspace{0.2cm}\noindent\textbf{Acknowledgements.} This work was funded by the project ANR TSIA DEVIN (ANR-23-IAS2-0001), ANER MOVIS from ``Region Bourgogne-Franche-Comte'' and ANR MANYVIS (ANR-23-CE23-0003-01), to whom we are grateful. 
We thank the access to the HPC computational resources of IDRIS under the allocation AD011016829.

\clearpage
\bibliographystyle{splncs04}
\bibliography{main}
\clearpage
\renewcommand{\thesection}{\Alph{section}}
\renewcommand{\thefigure}{S\arabic{figure}}
\renewcommand{\thetable}{S\arabic{table}}

\appendix

\begin{center}
    {\Large \textbf{[Supplementary Material] -- \\ \vspace{0.3cm} SVI360: Spherical Video Interpolation}}
\end{center}

\section{Additional Qualitative Visual Results}
\begin{figure*}[!htbp]
  \centering
  \scriptsize
  \setlength{\tabcolsep}{1.5pt}
  \renewcommand{\arraystretch}{1.0}
  \newcommand{\flowhalfbottomcrop}[1]{%
    \includegraphics[width=0.215\textwidth,height=0.185\textwidth,trim=200 0 0 256,clip]{#1}%
  }
  \newcommand{\flowmetriccellb}[3]{%
    \begin{tabular}{c}
      \flowhalfbottomcrop{#1} \\
      {\tiny PSNR #2, SSIM #3}
    \end{tabular}
  }
  \newcommand{\flowgtcellb}[1]{%
    \begin{tabular}{c}
      \flowhalfbottomcrop{#1}
    \end{tabular}
  }
  \begin{tabular}{cccc}
    \textbf{Method} & \textbf{im3} & \textbf{im5} & \textbf{im7} \\
    \textbf{Groundtruth} &
    \flowgtcellb{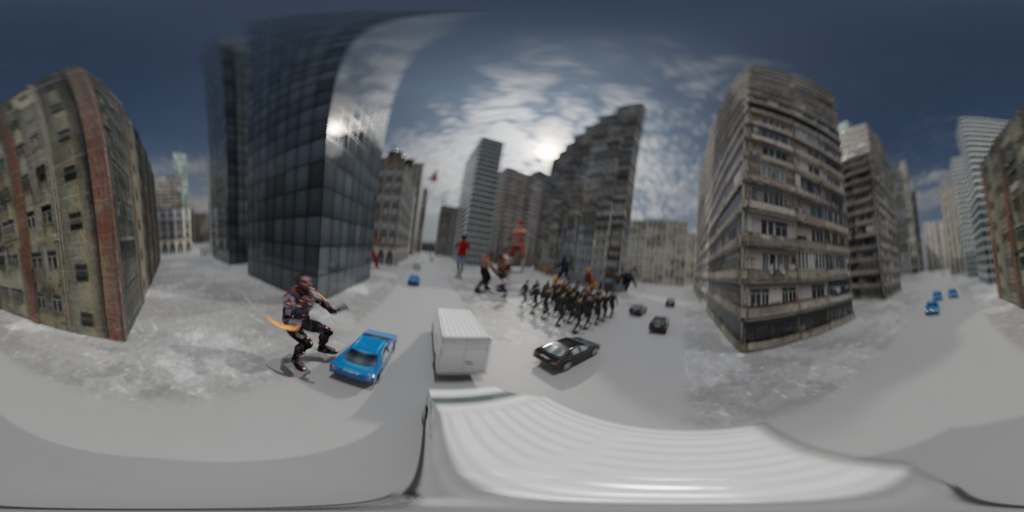} &
    \flowgtcellb{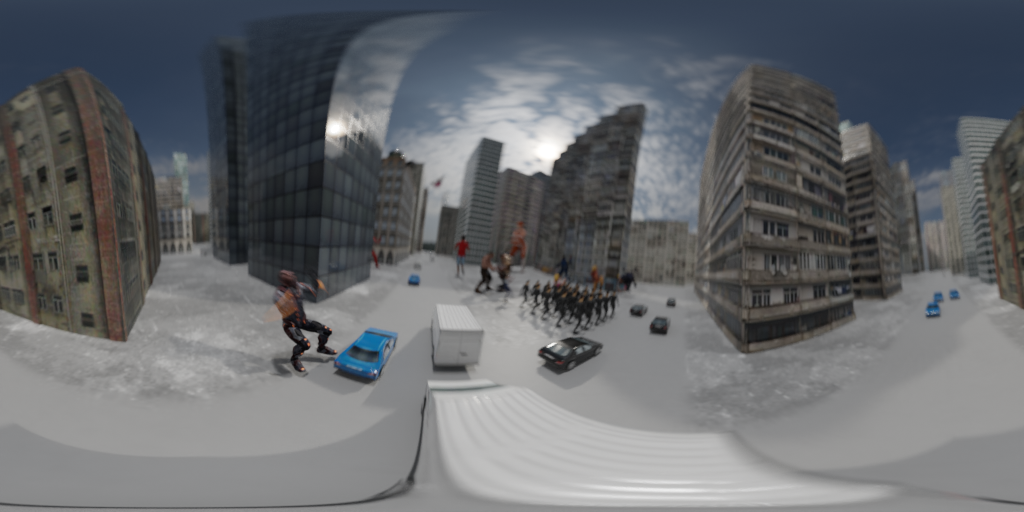} &
    \flowgtcellb{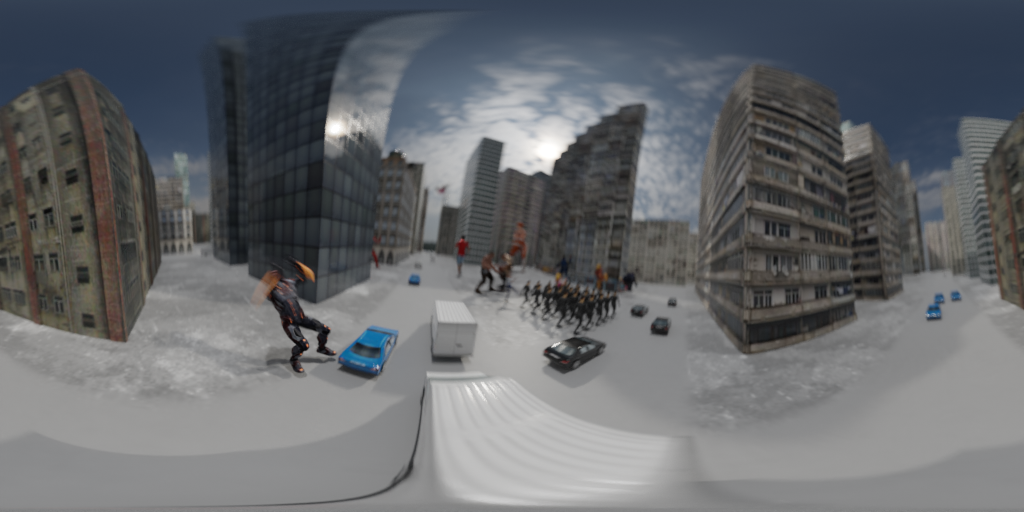} \\
    \textbf{SVI360 (ours)} &
    \flowmetriccellb{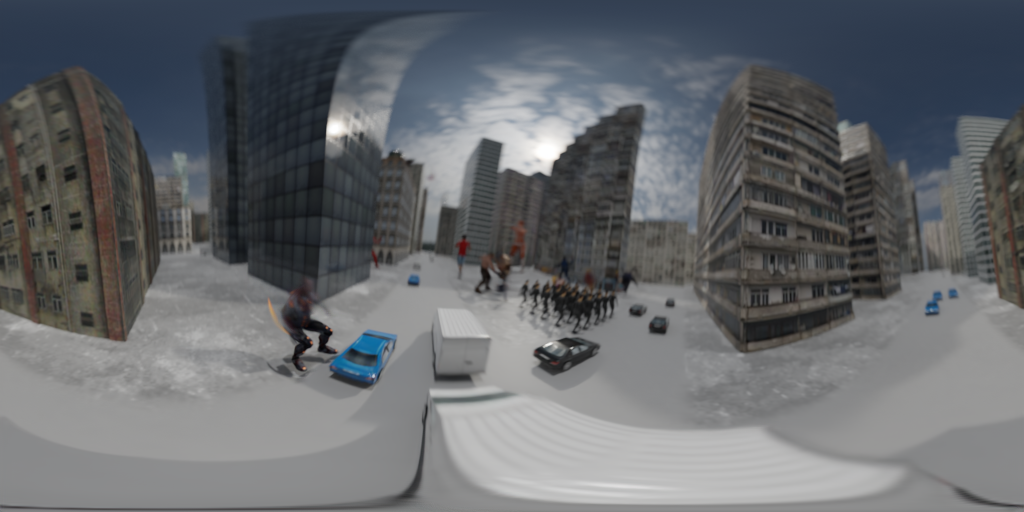}{36.94}{0.9828} &
    \flowmetriccellb{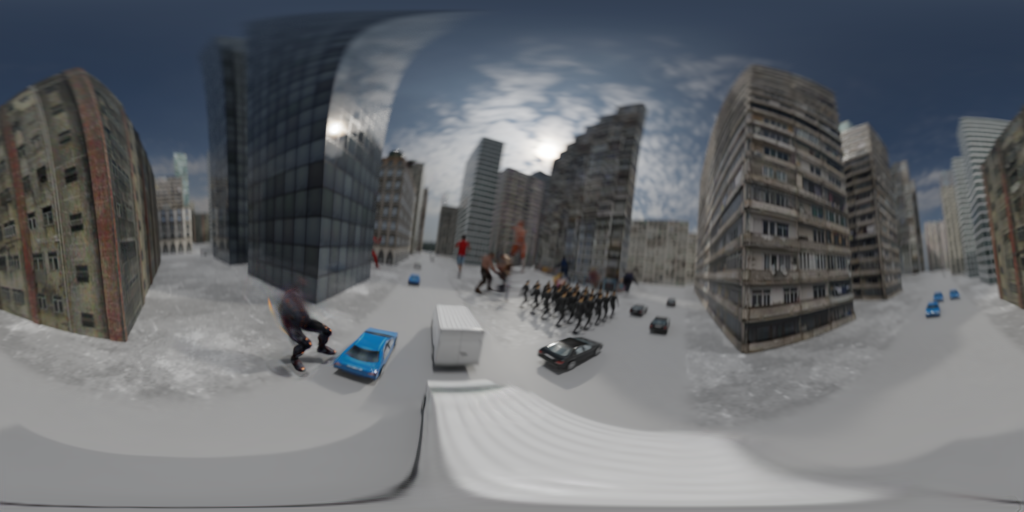}{37.12}{0.9801} &
    \flowmetriccellb{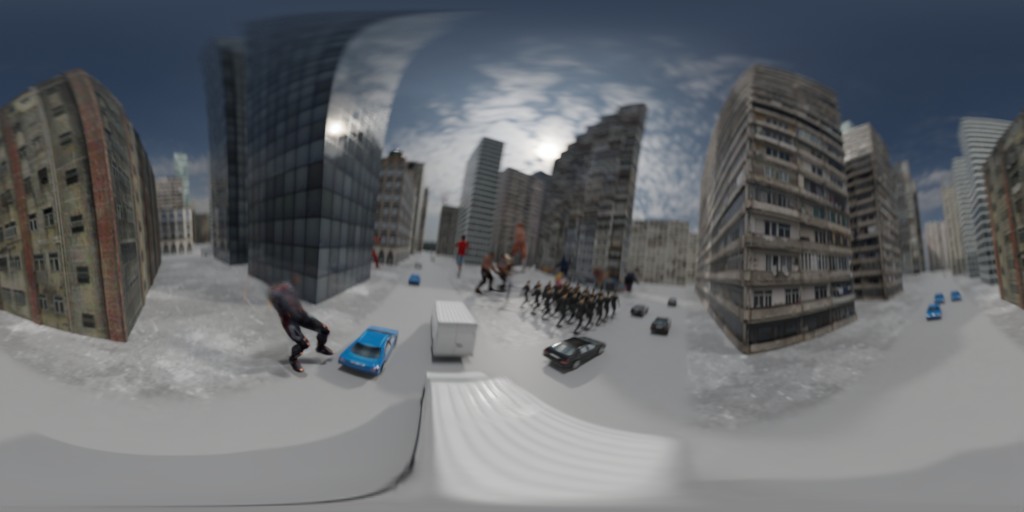}{37.90}{0.9840} \\
    \textbf{AMT-G~\cite{amt}} &
    \flowmetriccellb{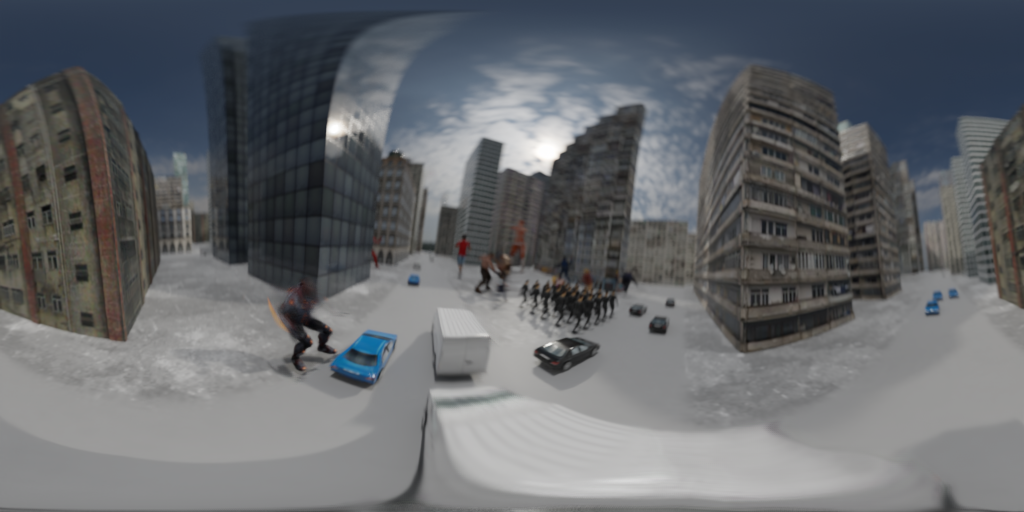}{31.84}{0.9627} &
    \flowmetriccellb{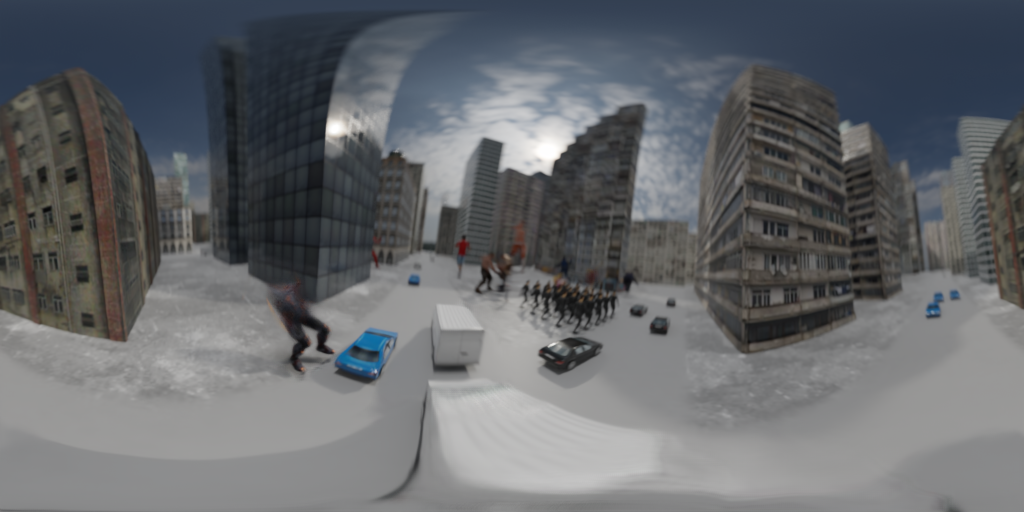}{30.27}{0.9615} &
    \flowmetriccellb{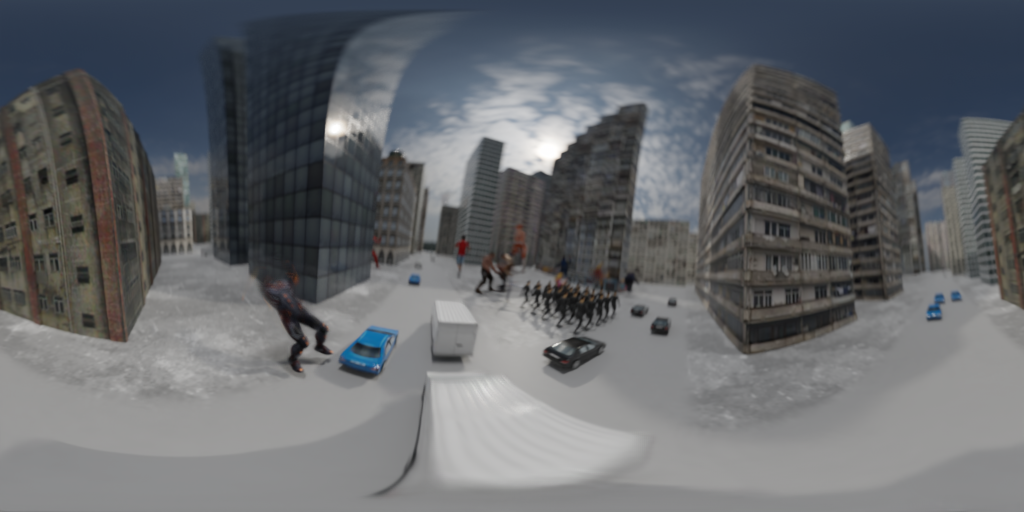}{31.66}{0.9711} \\
    \textbf{IFRNet-L~\cite{ifrnet}} &
    \flowmetriccellb{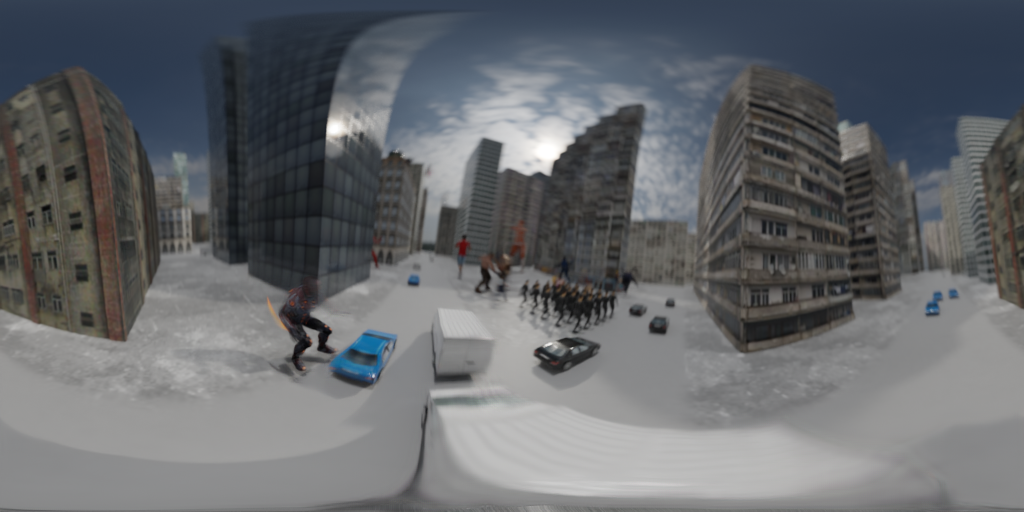}{31.23}{0.9578} &
    \flowmetriccellb{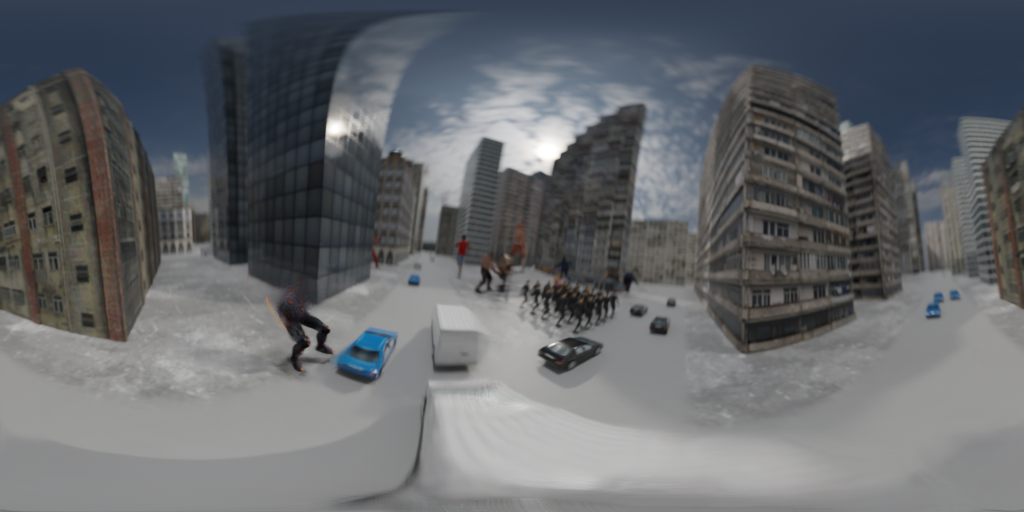}{29.97}{0.9562} &
    \flowmetriccellb{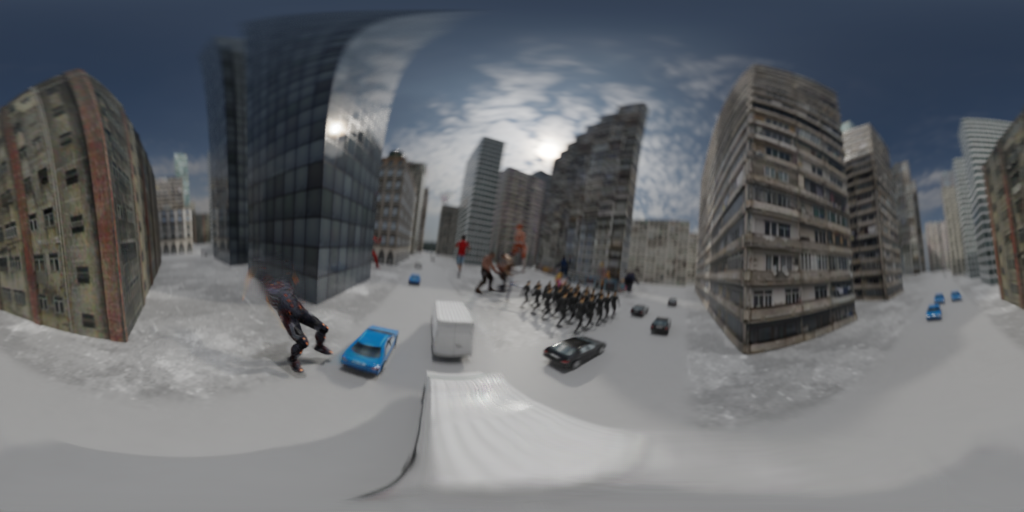}{31.60}{0.9672} \\
    \textbf{RIFE~\cite{rife}} &
    \flowmetriccellb{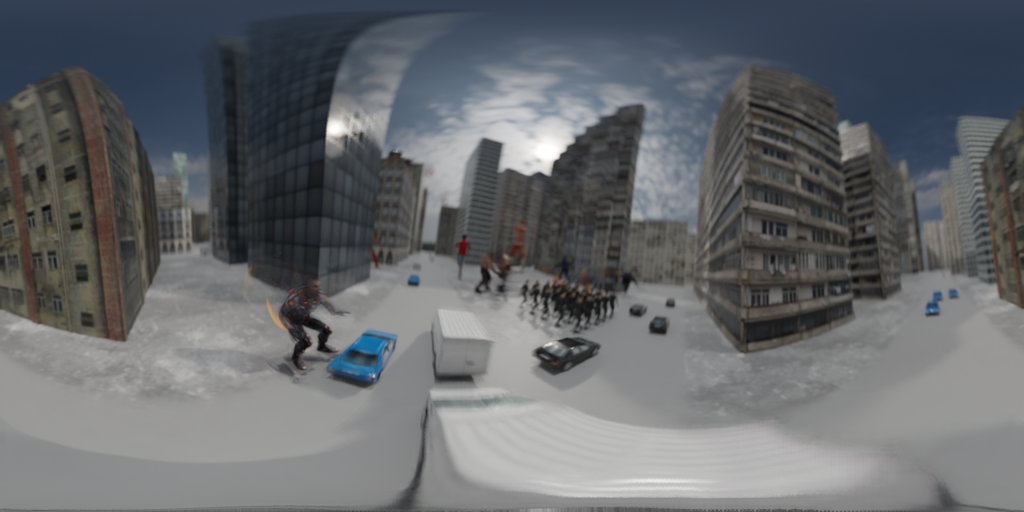}{31.07}{0.9511} &
    \flowmetriccellb{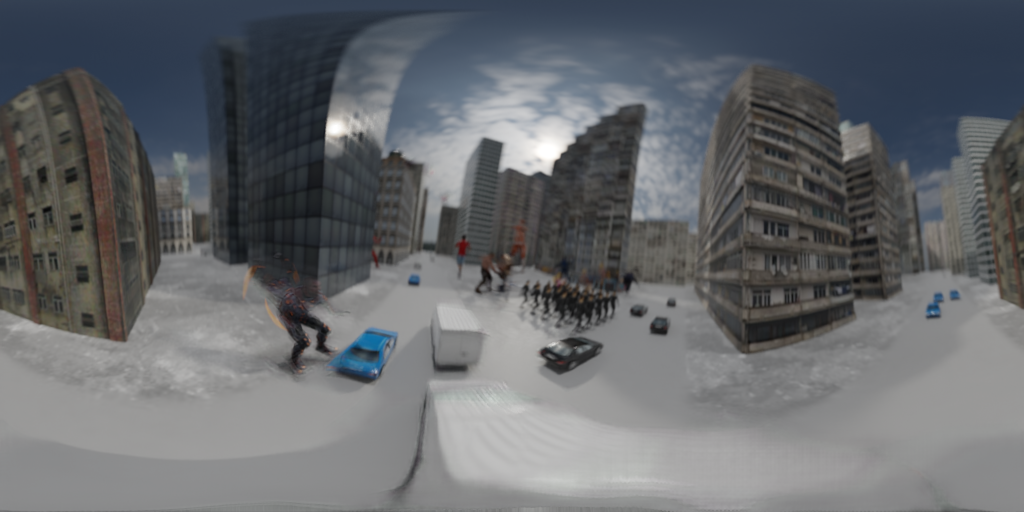}{30.48}{0.9509} &
    \flowmetriccellb{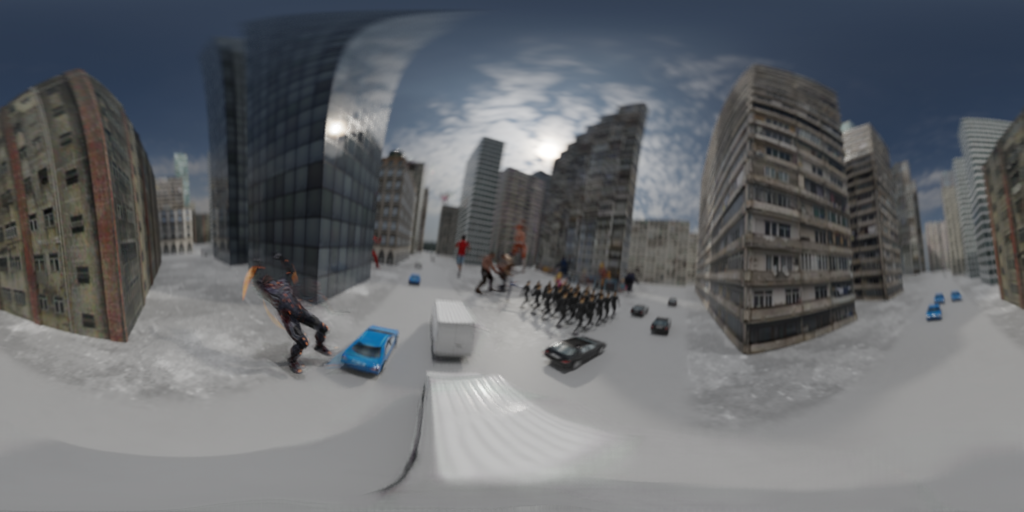}{31.19}{0.9633} \\
    \textbf{VFIFormer~\cite{vfiformer}} &
    \flowmetriccellb{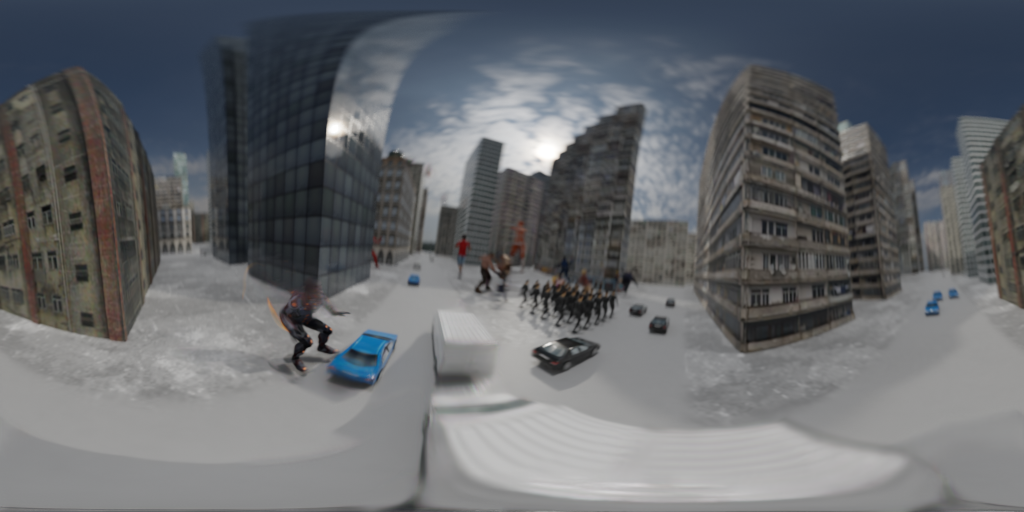}{30.43}{0.9495} &
    \flowmetriccellb{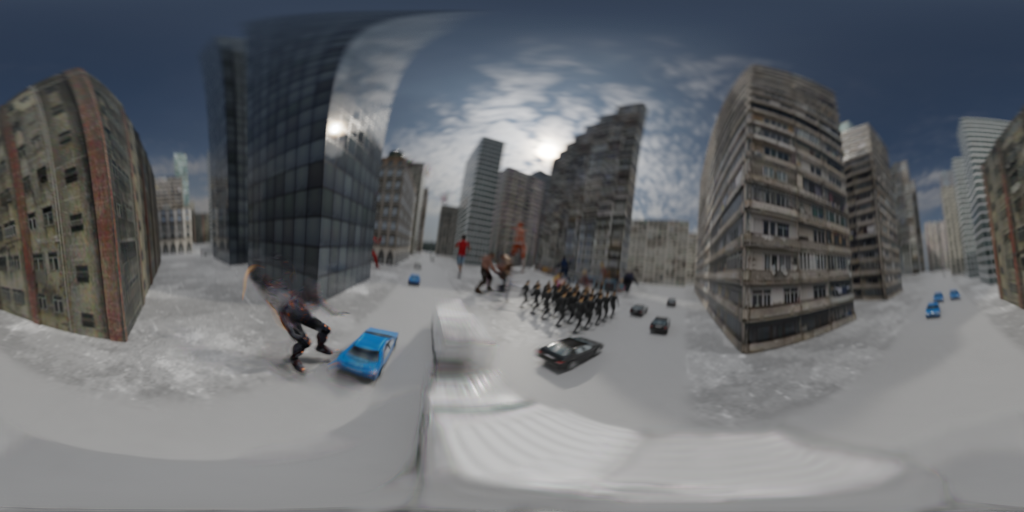}{29.55}{0.9445} &
    \flowmetriccellb{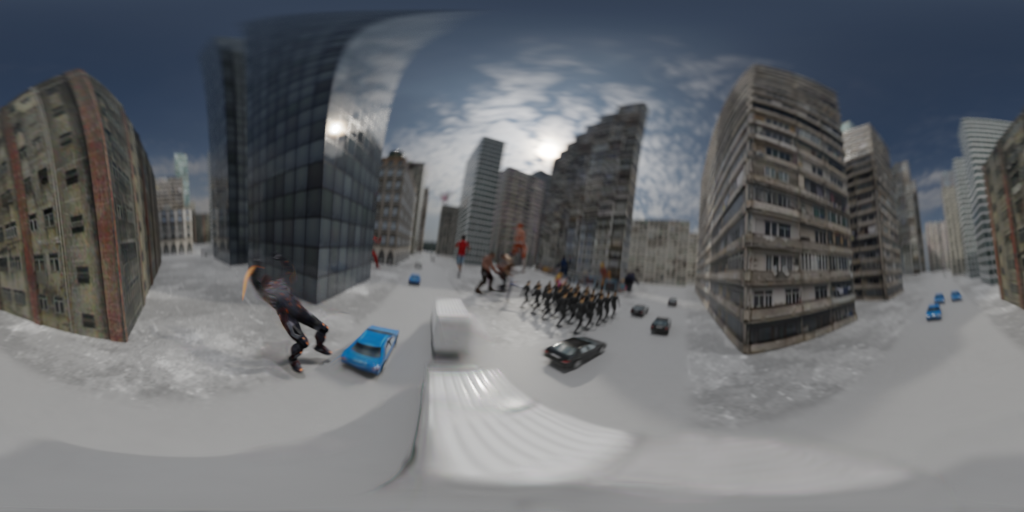}{30.46}{0.9602} \\
    \textbf{FCVG~\cite{FCVG}} &
    \flowmetriccellb{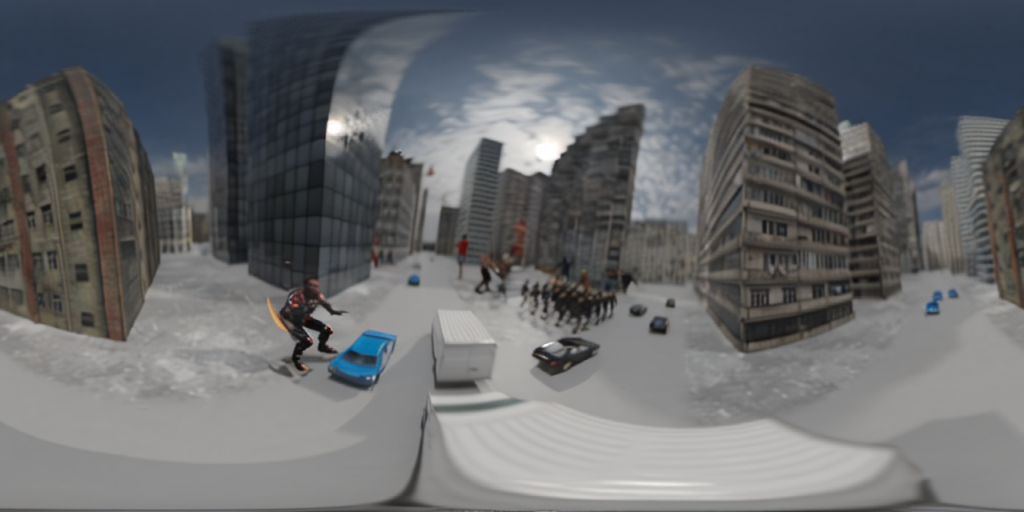}{29.44}{0.9203} &
    \flowmetriccellb{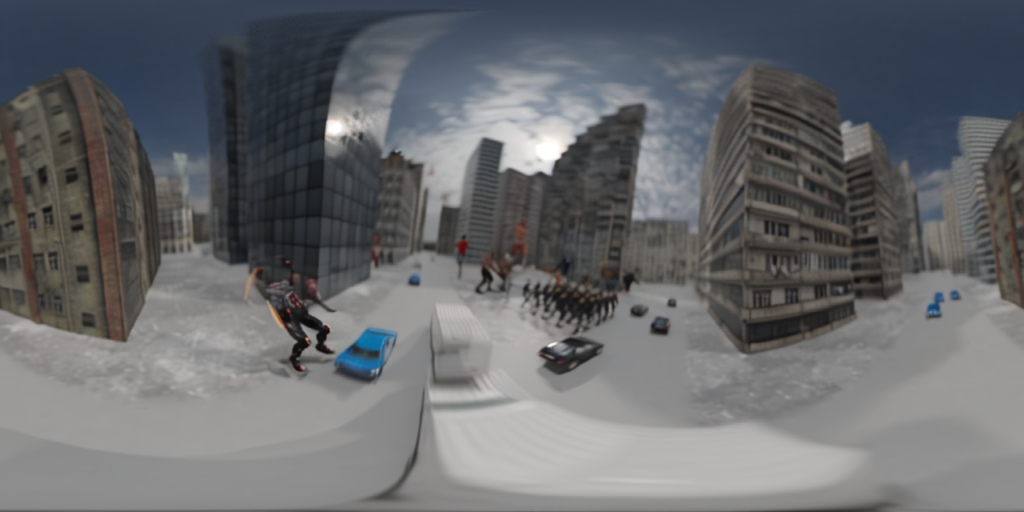}{28.81}{0.9275} &
    \flowmetriccellb{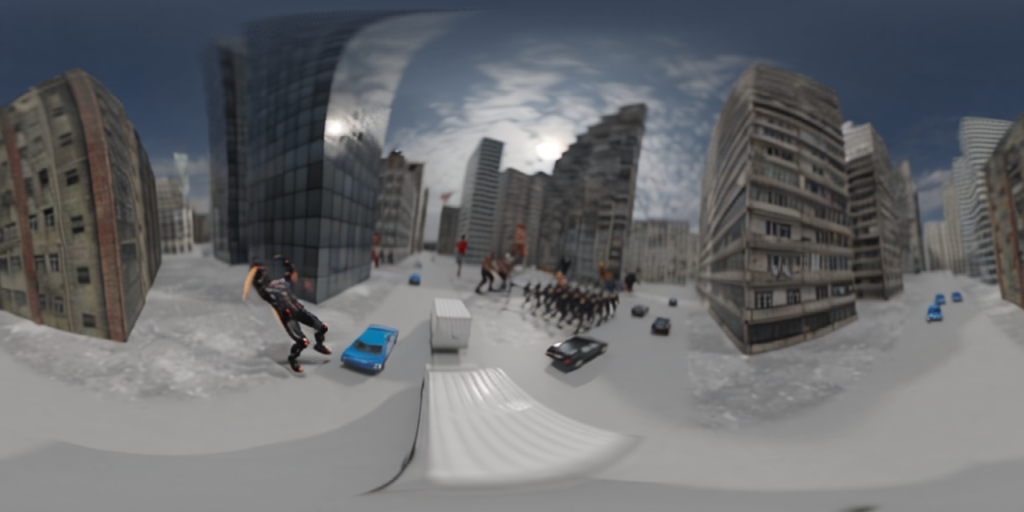}{29.39}{0.9394} \\
  \end{tabular}
  \caption{Additional qualitative results on Flow360 at three timesteps ($t=0.25$, $0.5$, and $0.75$). This scene contains a truck crossing the pole region, which causes severe distortion. Our method remains robust and produces reliable predictions even for arbitrary-time interpolation.}
  \label{fig:appendix_flow360_more_results}
\end{figure*}

In this section, we present additional visual results on two benchmark datasets, ODV360 and Flow360, to further show the interpolation improvements of the proposed SVI360 as shown in \cref{fig:appendix_odv360_challenging,fig:appendix_odv360_center_zoom,fig:appendix_flow360_more_results}. The
comparison methods include AMT-G~\cite{amt}, IFRNet-L~\cite{ifrnet},
RIFE~\cite{rife}, VFIFormer~\cite{vfiformer} and a generative method FCVG~\cite{FCVG}. For the Flow360 dataset, we take a nonuplet of frames as input and predict seven intermediate frames at timestamps 0.125, 0.25, ..., 0.875. For visualization, we present the interpolated frames at timestamps 0.25, 0.5, and 0.75 as representative examples. Please also see the interpolations of videos provided in the project website: \url{https://icb-vision-ai.github.io/video360_interpolation/}
\newcommand{\odvcell}[2]{\shortstack{\textbf{#1}\\#2}}
\newcommand{\odvrow}[6]{%
  \odvcell{#1}{#2} &
  \odvcell{#3}{#4} &
  \odvcell{#5}{#6} \\
}

\begin{figure*}[h]
  \centering
  \scriptsize
  \setlength{\tabcolsep}{2pt}
  \renewcommand{\arraystretch}{1.0}
  \newcommand{\odvfullimg}[1]{\includegraphics[width=0.30\textwidth]{#1}}
  \begin{tabular}{ccc}
    \multicolumn{3}{c}{\textbf{Overlaid}} \\
    \multicolumn{3}{c}{\includegraphics[width=0.62\textwidth]{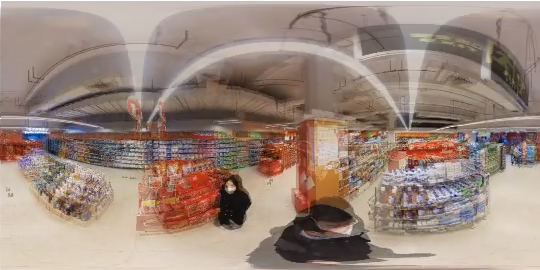}} \\
    \odvrow
      {Groundtruth}{\odvfullimg{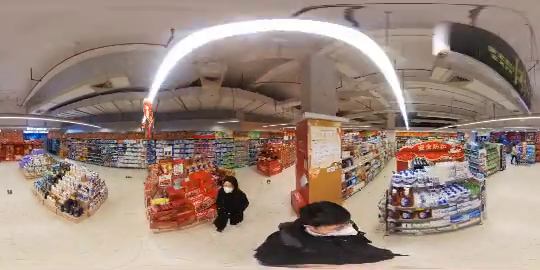}}
      {SVI360 (Ours)}{\odvfullimg{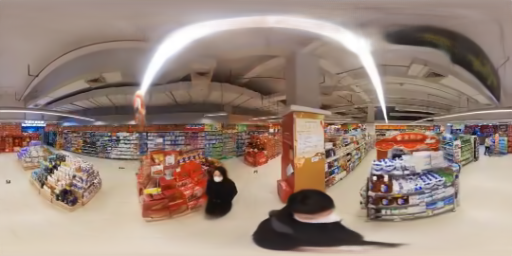}}
      {AMT-G~\cite{amt}}{\odvfullimg{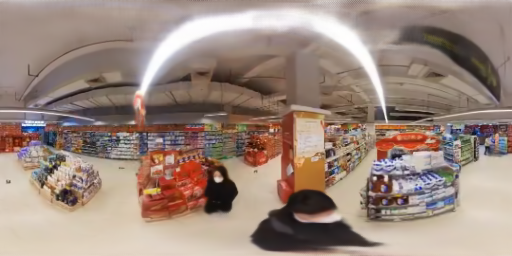}}
    \odvrow
      {IFRNet-L~\cite{ifrnet}}{\odvfullimg{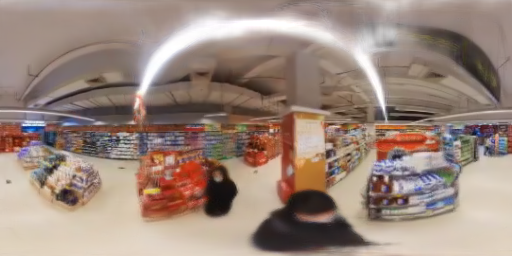}}
      {RIFE~\cite{rife}}{\odvfullimg{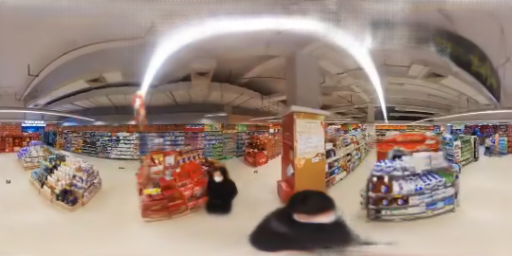}}
      {VFIFormer~\cite{vfiformer}}{\odvfullimg{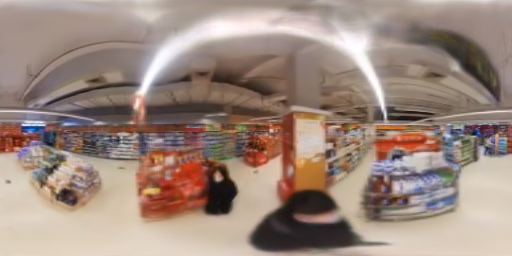}}
  \end{tabular}
  \caption{Challenging interpolation case with large camera displacement, where the light bulb passes through the pole region and the two input frames differ significantly (as it can be noticed in the overlaid frames). Even in these conditions, our method still preserves details in highly distorted areas.}
  \label{fig:appendix_odv360_challenging}
\end{figure*}

\begin{figure*}[!t]
  \centering
  \scriptsize
  \setlength{\tabcolsep}{2pt}
  \renewcommand{\arraystretch}{1.0}
  \newcommand{\odvfullimgone}[1]{\includegraphics[width=0.30\textwidth,trim=0 75 300 90,clip]{#1}}
  \begin{tabular}{ccc}
    \multicolumn{3}{c}{\textbf{Overlaid}} \\
    \multicolumn{3}{c}{\includegraphics[width=0.62\textwidth, trim=0 75 300 90,clip]{figures/odv_results_1/GT_overlay.png}} \\
    \odvrow
      {Groundtruth}{\odvfullimgone{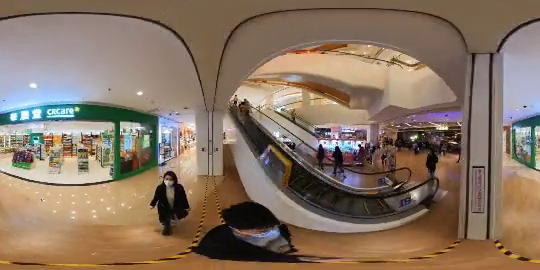}}
      {SVI360 (Ours)}{\odvfullimgone{figures/odv_results_1/SVI360_PSNR_26.58_SSIM_0.9292.png}}
      {AMT-G~\cite{amt}}{\odvfullimgone{figures/odv_results_1/AMT_PSNR_26.01_SSIM_0.9207.png}}
    \odvrow
      {IFRNet-L~\cite{ifrnet}}{\odvfullimgone{figures/odv_results_1/IFRNet_PSNR_25.02_SSIM_0.8938.png}}
      {RIFE~\cite{rife}}{\odvfullimgone{figures/odv_results_1/RIFE_PSNR_25.05_SSIM_0.8993.png}}
      {VFIFormer~\cite{vfiformer}}{\odvfullimgone{figures/odv_results_1/VFIformer_PSNR_24.81_0.8881.png}}
  \end{tabular}
  \caption{This example shows the capability of our approach to also preserve fine details near the central region of the spherical image. Please notice the store name text remains recognizable (comparable to AMT-G), while the results from all other methods are noticeably blurrier.}
  \label{fig:appendix_odv360_center_zoom}
\end{figure*}

\section{Dataset and Augmentation Details}

\noindent\textbf{Dataset details:}
\begin{itemize}
  \item \textbf{FlowScape}: training and inference used input of shape 1024$\times$512.
  \item \textbf{Flow360}: training and inference used input of shape 1024$\times$512. The dataset is organized into 9-frame tuples. During training, the first and last frames are used to interpolate one randomly selected intermediate frame among the seven inner frames. During evaluation, all seven intermediate frames are interpolated and compared against ground truth.
  \item \textbf{ODV360}: the original resolution is 540$\times$270, but for both training and inference we resize frames to 512$\times$256 so that models with up to 8$\times$ spatial downsampling do not suffer from shape-mismatch issues.
  \item \textbf{360VFI}: training and inference used input of shape 480$\times$240.
\end{itemize}

\noindent\textbf{Augmentation details.}
Our implementation applies the following augmentations, where $p$ denotes the probability of applying each method:
random rotation roll/pitch/yaw ($p=0.3$, rotation ratio $=0.1$),
reverse channel ($p=0.5$),
color jitter ($p=0.3$),
vertical flip ($p=0.3$),
horizontal flip ($p=0.3$),
temporal reversal ($p=0.5$), and
random erasing ($p=0.3$).

\section{Spherical Weighted Loss}
We analyze the contribution of the spherical weighted Charbonnier loss for our SVI360 model and the strong SOTA baseline AMT-G~\cite{amt}. The unified comparison in
\cref{tab:supp_spherical_loss_comparison} shows that the spherical loss improves the spherical metrics of SVI360, increasing WS-PSNR from 37.63 to 37.73 and WS-SSIM from
0.9638 to 0.9647. This gain comes with a small trade-off in the conventional PSNR and SSIM, which is expected because the loss emphasizes the perceptually and geometrically
important regions of the equirectangular image rather than treating all pixels equally.

The same trend is observed when the loss is applied to AMT-G. Compared with the standard AMT-G training objective, AMT-G with the spherical weighted loss improves from
38.05/0.9704 to 38.29/0.9738 in PSNR/SSIM and from 37.40/0.9622 to 37.42/0.9629 in WS-PSNR/WS-SSIM. These results confirm that the loss itself is beneficial for spherical video interpolation. However, AMT-G with the proposed loss still remains below SVI360, showing that the final performance also depends on the
dual-branch refinement and fusion design. A representative qualitative example is
shown in \cref{fig:ablation_spherical_loss}, where the spherical loss better preserves
center-region details that contain major visual information. Additional central and polar
crops in \cref{fig:supp_amtg_spherical_loss} show the same behavior for AMT-G with and
without the spherical loss.

\begin{table}[tb]
  \centering
  \small
  \setlength{\tabcolsep}{4pt}
  \caption{Unified comparison of AMT-G and SVI360 with/without spherical weighted
  loss, plus LiteSVI360, on FlowScape (1024$\times$512).}
  \label{tab:supp_spherical_loss_comparison}
  \resizebox{\linewidth}{!}{%
  \begin{tabular}{@{}llccccc@{}}
    \toprule
    Method & Sph. loss & PSNR & WS-PSNR & SSIM & WS-SSIM & Params (M) \\
    \midrule
    AMT-G~\cite{amt} & No & 38.05 & 37.40 & 0.9704 & 0.9622 & 30.6 \\
    AMT-G~\cite{amt} & Yes & 38.29 & 37.42 & 0.9738 & 0.9629 & 30.6 \\
    SVI360 & No & \textbf{38.72} & 37.63 & \textbf{0.9761} & 0.9638 & 50.1 \\
    LiteSVI360 & Yes & 38.55 & 37.63 & 0.9744 & 0.9641 & 22.9 \\
    SVI360 & Yes & 38.68 & \textbf{37.73} & 0.9754 & \textbf{0.9647} & 50.1 \\
    \bottomrule
  \end{tabular}%
  }
\end{table}

\begin{figure}[!t]
  \centering
  \scriptsize
  \setlength{\tabcolsep}{2pt}
  \begin{tabular}{ccccc}
    & GT & AMT-G & AMT-G+sph. & SVI360 \\
    \rotatebox[origin=c]{90}{Central} &
    \includegraphics[width=0.19\linewidth]{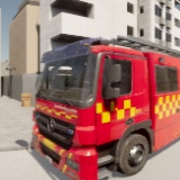} &
    \includegraphics[width=0.19\linewidth]{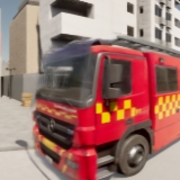} &
    \includegraphics[width=0.19\linewidth]{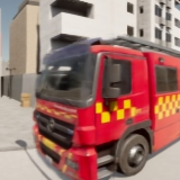} &
    \includegraphics[width=0.19\linewidth]{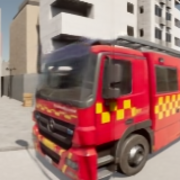} \\
    \rotatebox[origin=c]{90}{Pole} &
    \includegraphics[width=0.19\linewidth]{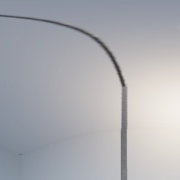} &
    \includegraphics[width=0.19\linewidth]{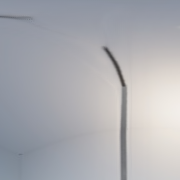} &
    \includegraphics[width=0.19\linewidth]{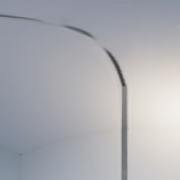} &
    \includegraphics[width=0.19\linewidth]{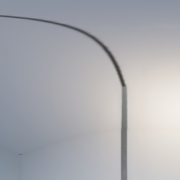} \\
  \end{tabular}
  \caption{Qualitative comparison of AMT-G trained with standard and spherical
  weighted losses against SVI360. The top row shows a central-region crop, and the
  bottom row shows a polar-region crop.}
  \label{fig:supp_amtg_spherical_loss}
\end{figure}

\section{LiteSVI360}
We also report results with a reduced version of our model (LiteSVI360), a smaller variant that uses AMT-L for the primitive branch
and AMT-S for the orthogonal branch instead of the AMT-G+AMT-L combination used in
SVI360. LiteSVI360 has only 22.9M parameters, compared with 30.6M for AMT-G and
50.1M for SVI360, yet it still outperforms both AMT-G and AMT-G trained with the
spherical loss, as shown in \cref{tab:supp_spherical_loss_comparison}. This result shows that the
performance gain does not come only from increasing the number of parameters.

\section{Comparison with Generative Inbetweening}
We also compare with FCVG~\cite{FCVG}, a recent generative inbetweening method, on the Flow360 nonuplet dataset. Since FCVG is designed to generate plausible intermediate frames rather than strictly reconstruct pixel-aligned ground truth, the reported metrics should be interpreted with caution. Metrics such as PSNR, SSIM, WS-PSNR, and WS-SSIM favor deterministic methods that produce geometrically faithful reconstructions, while they may penalize generative outputs that are perceptually plausible but not exactly aligned with the reference frame.

\begin{figure*}[tb]
  \centering
  \scriptsize
  \setlength{\tabcolsep}{1pt}
  \renewcommand{\arraystretch}{0.85}
  \begin{tabular}{cccc}
    \textbf{Frame} & \textbf{FCVG} & \textbf{SVI360 (ours)} & \textbf{AMT-G} \\
    im1 & \includegraphics[width=0.31\textwidth]{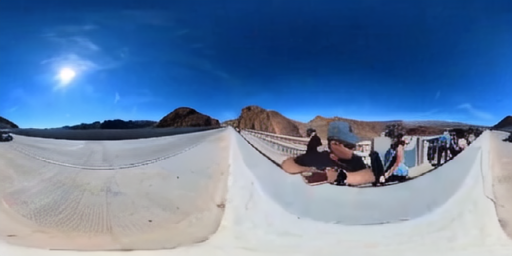} & \includegraphics[width=0.31\textwidth]{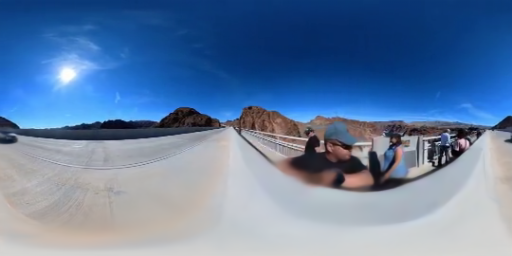} & \includegraphics[width=0.31\textwidth]{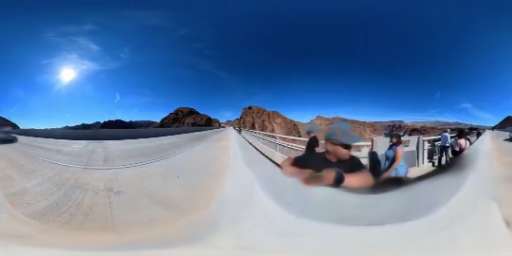} \\
    im2 & \includegraphics[width=0.31\textwidth]{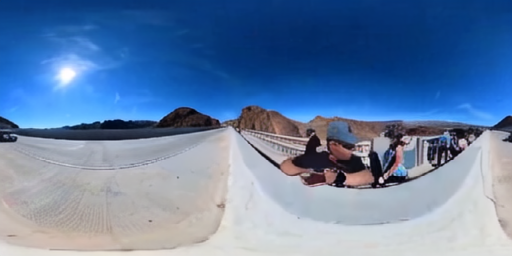} & \includegraphics[width=0.31\textwidth]{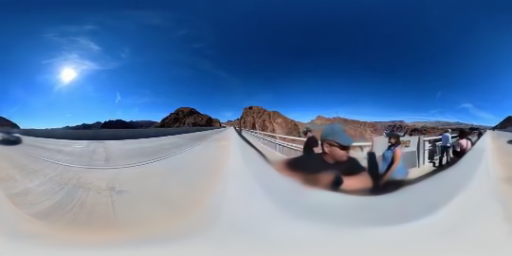} & \includegraphics[width=0.31\textwidth]{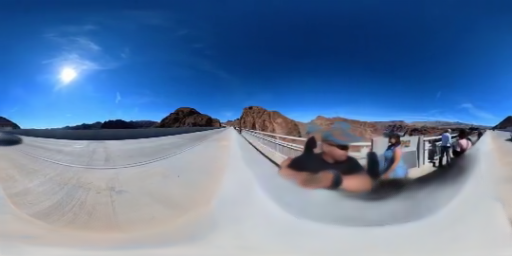} \\
    im3 & \includegraphics[width=0.31\textwidth]{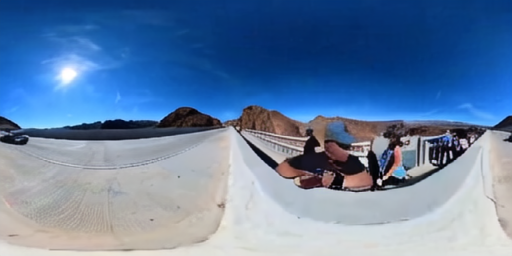} & \includegraphics[width=0.31\textwidth]{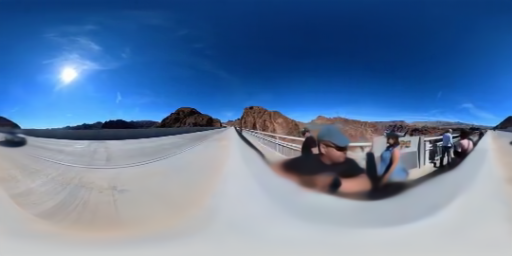} & \includegraphics[width=0.31\textwidth]{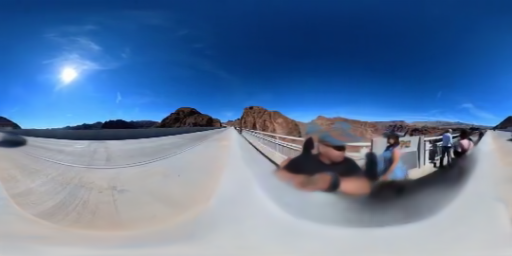} \\
    im4 & \includegraphics[width=0.31\textwidth]{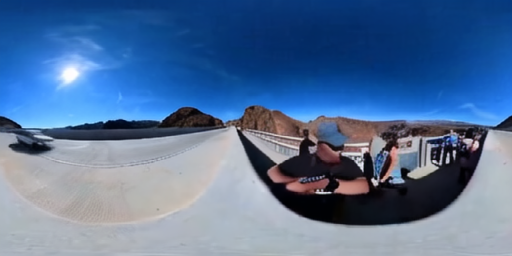} & \includegraphics[width=0.31\textwidth]{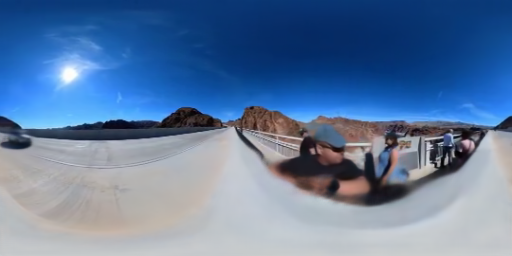} & \includegraphics[width=0.31\textwidth]{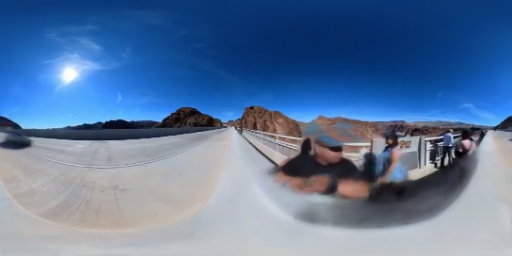} \\
    im5 & \includegraphics[width=0.31\textwidth]{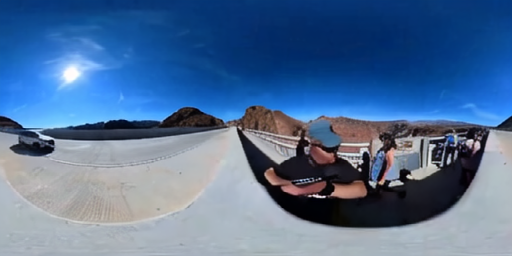} & \includegraphics[width=0.31\textwidth]{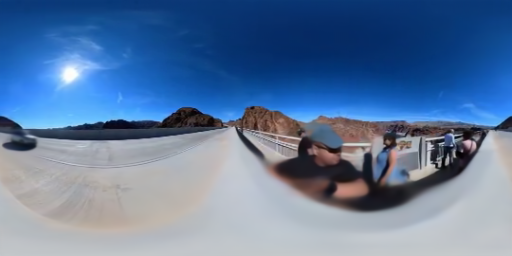} & \includegraphics[width=0.31\textwidth]{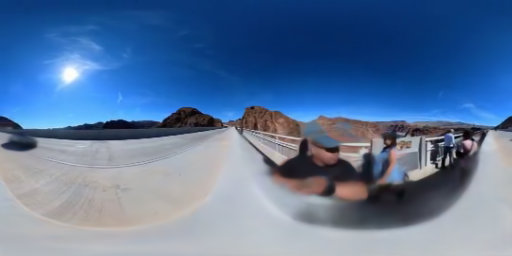} \\
    im6 & \includegraphics[width=0.31\textwidth]{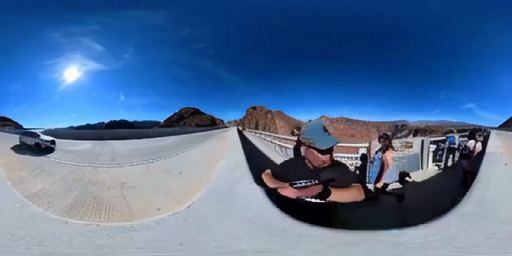} & \includegraphics[width=0.31\textwidth]{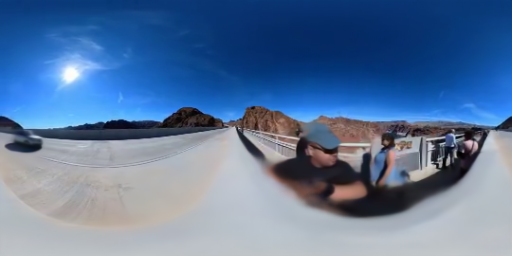} & \includegraphics[width=0.31\textwidth]{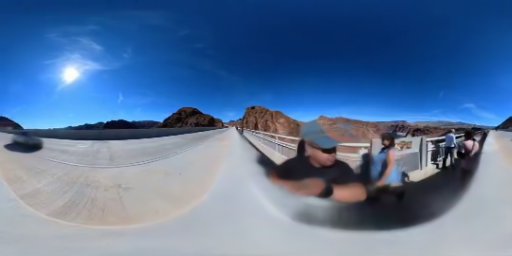} \\
    im7 & \includegraphics[width=0.31\textwidth]{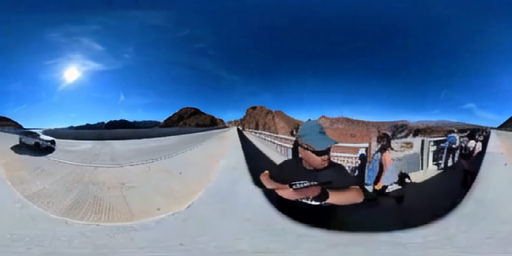} & \includegraphics[width=0.31\textwidth]{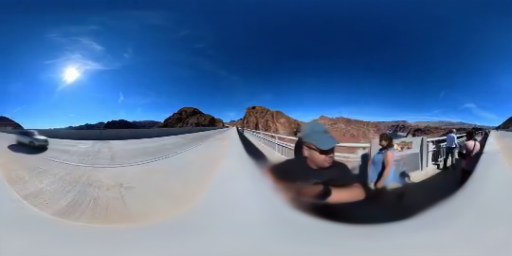} & \includegraphics[width=0.31\textwidth]{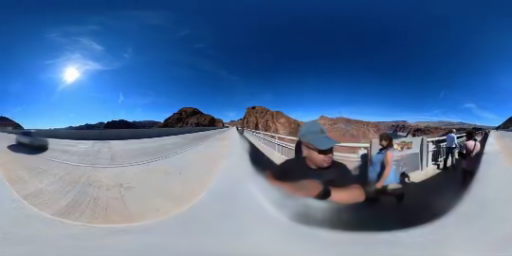} \\
  \end{tabular}
  \caption{Qualitative comparison among FCVG, AMT-G, and SVI360 on 360VFI Extreme Setting Scene. FCVG preserves well the local details but shows less accurate temporal motion, with the car position changing abruptly between im3 and im4, whereas SVI360 maintains smoother and more consistent motion across the sequence.}
  \label{fig:supp_fcvg_svi360_qualitative}
\end{figure*}

Under this evaluation protocol, Tab.~\ref{tab:supp_fcvg_flow360} shows that FCVG
remains below SVI360 on all reported spherical interpolation metrics. This indicates
that SVI360 is more suitable when accurate spherical frame reconstruction is required. However, FCVG can still produce good local structures under large motion, and generative inbetweening remains promising for real world scenarios where the intermediate content is ambiguous. Qualitative results on Flow360 are shown in ~\cref{fig:appendix_flow360_more_results}. The qualitative comparison ~\cref{fig:supp_fcvg_svi360_qualitative} further highlights the different behavior of the two
paradigms. FCVG preserves sharp local appearance and fine details, while
flow-based methods such as SVI360 and AMT-G can produce blurrier textures in
some regions. However, FCVG does not always maintain accurate temporal motion:
for example, the car position changes abruptly from im3 to im4 in ~\cref{fig:supp_fcvg_svi360_qualitative}. In contrast, SVI360 produces smoother and more geometrically consistent motion across the interpolated sequence.

\begin{table}[tb]
  \centering
  \caption{FCVG and SVI360 comparison on the Flow360 dataset.}
  \label{tab:supp_fcvg_flow360}
  \begin{tabular}{lcccc}
    \toprule
    Method & PSNR & WS-PSNR & SSIM & WS-SSIM \\
    \midrule
    FCVG & 28.28 & 27.50 & 0.8953 & 0.8768 \\
    SVI360 & \textbf{34.02} & \textbf{33.30} & \textbf{0.9752} & \textbf{0.9615} \\
    \bottomrule
  \end{tabular}
\end{table}

\section{Architecture Details}
\begin{figure*}[!htbp]
  \centering
  \includegraphics[width=0.95\textwidth]{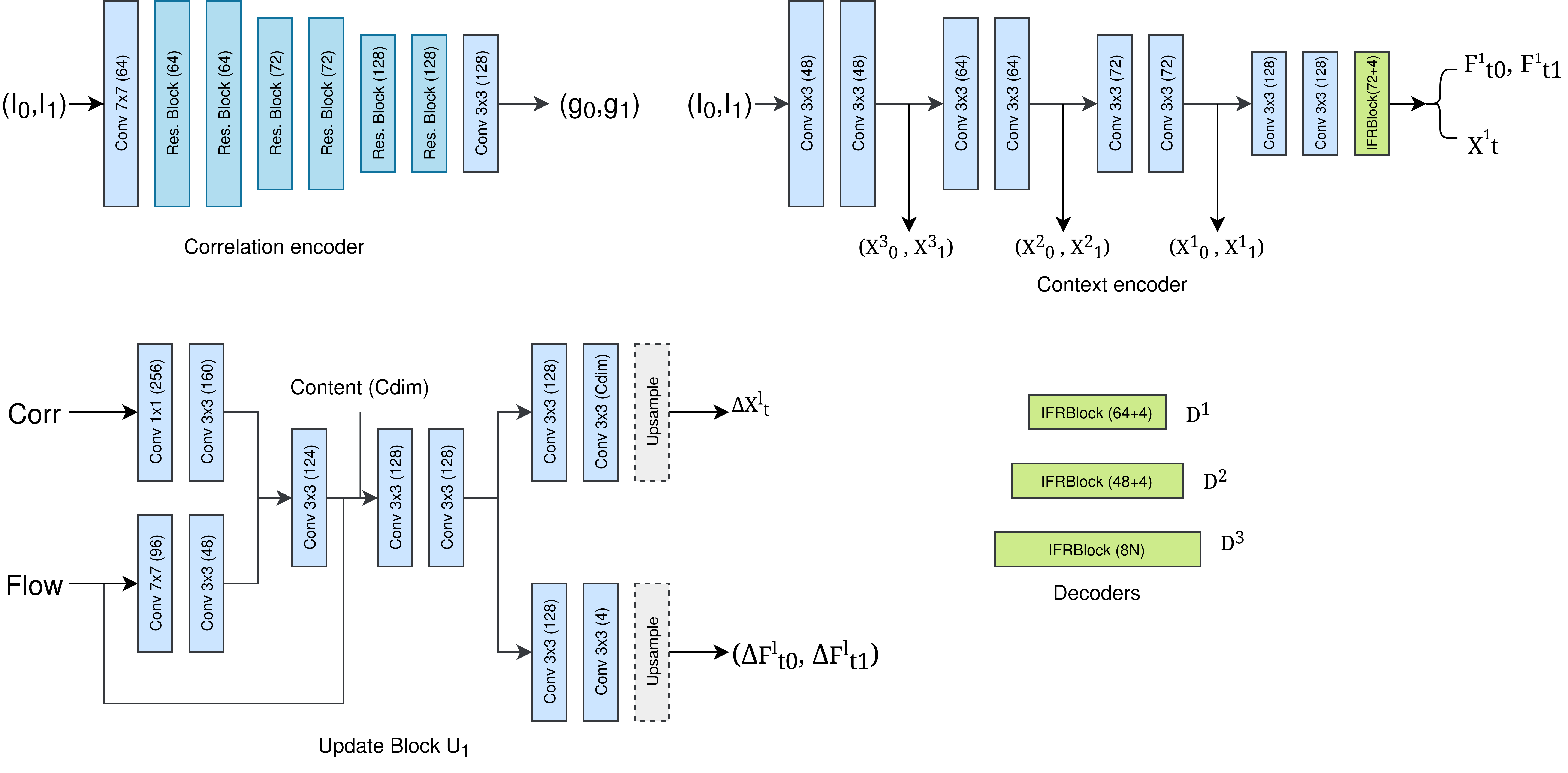}
  \caption{Orthogonal-branch architecture details.}
  \label{fig:appendix_ortho_branch}
\end{figure*}

\begin{figure*}[!htbp]
  \centering
  \includegraphics[width=0.95\textwidth]{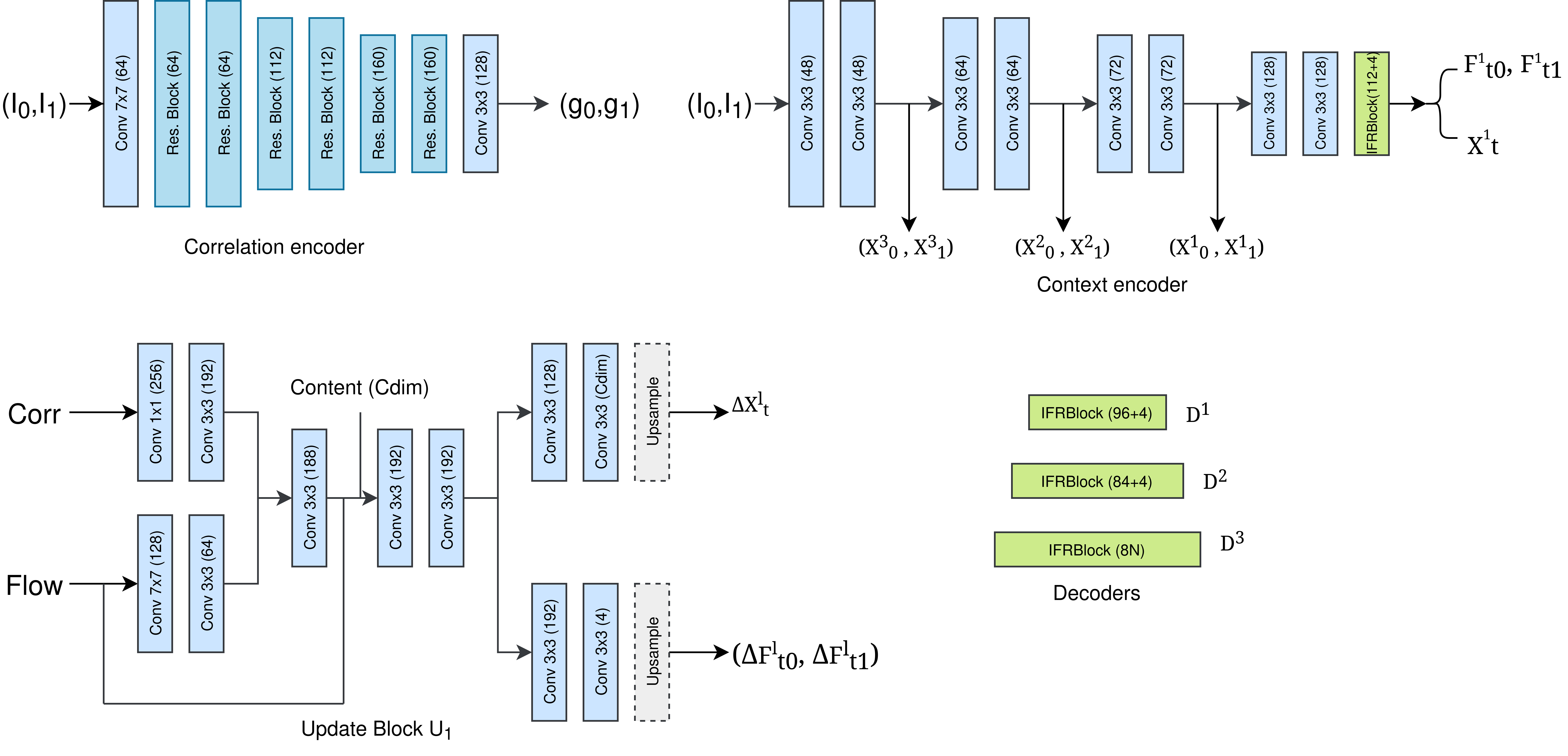}
  \caption{Primitive-branch architecture details.}
  \label{fig:appendix_primitive_branch}
\end{figure*}

\begin{figure*}[!htbp]
  \centering
  \includegraphics[width=0.8\textwidth]{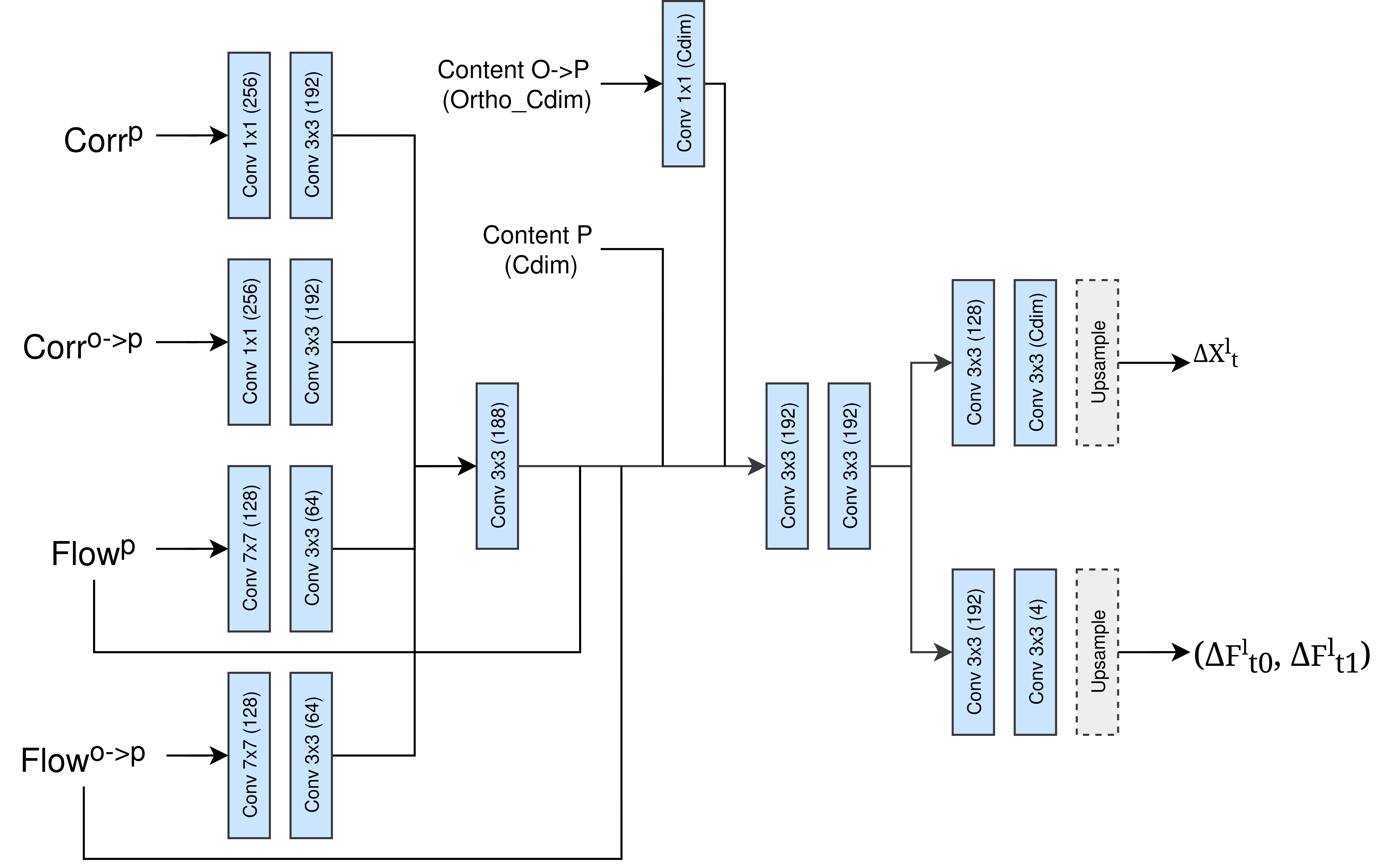}
  \caption{Detailed structure of the update block $U_{\text{2}}$ used in the Spherical Refinement (SR block). The $1\times1$ convolution fed by Content O$\rightarrow$P is responsible for adapting the intermediate-feature dimension when the two branches use different architectures (and thus different feature dimensions).}
  \label{fig:appendix_update_block}
\end{figure*}

We build our model on top of the AMT framework~\cite{amt}. The orthogonal branch is
inspired by the AMT-L design, while the primitive branch follows the AMT-G design to
place more focus on primitive-image refinement, as shown in \cref{fig:appendix_ortho_branch} and \cref{fig:appendix_primitive_branch}. Compared with AMT architecture, our model introduces three major key modifications. First, we design a dual-cost collaborative lookup that queries two correlation volumes from two views, instead of using a single correlation volume. This design helps to model distortion near polar regions and provides more reliable refinement signals for
both branches. Second, we add an SR block to the primitive branch to absorb
complementary information from the orthogonal branch. This cross-view interaction
improves both optical-flow refinement and intermediate-feature updating in the
primitive branch. Third, unlike AMT, our training does not rely on auxiliary
optical-flow supervision. Instead, the model is optimized with a spherical weighted
reconstruction loss, which accounts for the non-uniform latitude sampling of
equirectangular images and encourages a fairer contribution of different spherical
regions during optimization.
Note that the update block $U_1$ in the orthogonal branch is different (and less
complex) than that in the primitive branch. The lookup radius is set to 3. The IFRBlock denotes the decoder proposed in IFRNet~\cite{ifrnet}, which
jointly estimates bilateral flows and the intermediate feature. The update block
$U_2$ in SR is shown in \cref{fig:appendix_update_block}; it is essentially the
same as $U_1$, but with an additional input.

\end{document}